\documentclass{article}

\usepackage[preprint]{neurips_2025}

\usepackage[utf8]{inputenc} % allow utf-8 input
\usepackage[T1]{fontenc}    % use 8-bit T1 fonts
\usepackage[backref=page]{hyperref}       % hyperlinks
\usepackage{url}            % simple URL typesetting
\usepackage{booktabs}       % professional-quality tables
\usepackage{amsfonts}       % blackboard math symbols
\usepackage{nicefrac}       % compact symbols for 1/2, etc.
\usepackage{microtype}      % microtypography
\usepackage{xcolor}         % colors

\usepackage{graphicx}
\usepackage{multirow}
\usepackage{mathtools}
\usepackage{caption}
\usepackage[shortlabels]{enumitem}
\usepackage{wrapfig}
\usepackage{subcaption}
\usepackage{tabularx}
\usepackage{dblfloatfix}
\usepackage{bm}
\usepackage{tcolorbox}

\tcbset{
  highlightquestion/.style={
    colframe=black,
    arc=2mm,
    left=1mm,
    right=1mm,
    colback=gray!10, 
  }
}
\newcommand{\Xmatrix}{\bm{X}}
\newcommand{\xvector}{\bm{x}}
\newcommand{\Wmatrix}{\bm{W}}

\newcommand{\Dmatrix}{\bm{D}}
\newcommand{\dvector}{\bm{d}}
\newcommand{\zvector}{\bm{z}}
\newcommand{\Zmatrix}{\bm{Z}}
\newcommand{\Qmatrix}{\bm{Q}}
\newcommand{\Kmatrix}{\bm{K}}
\newcommand{\muvector}{\bm{\mu}}

\newcommand{\eg}{{\it e.g.}}
\newcommand{\ie}{{\it i.e.}}

\newtheorem{theorem}{Theorem}[section]

\newtheorem{definition}[theorem]{Definition}
\newtheorem{assumption}[theorem]{Assumption}
\title{Hyper-SET: Designing Transformers via Hyperspherical Energy Minimization}

\author{%
  Yunzhe Hu \\
  School of Computing and Data Science\\
  The University of Hong Kong\\
  \texttt{yzhu@cs.hku.hk} \\
  \And
  Difan Zou \\
  School of Computing and Data Science \\
  The University of Hong Kong \\
  \texttt{dzou@cs.hku.hk} \\
  \And
  Dong Xu \\
  School of Computing and Data Science \\
  The University of Hong Kong \\
  \texttt{dongxu@cs.hku.hk} \\
}

\begin{document}

\maketitle

\begin{abstract}

Transformer-based models have achieved remarkable success, but their core components, Transformer layers, are largely heuristics-driven and engineered from the bottom up, calling for a prototypical model with high interpretability and practical competence. To this end, we conceptualize a principled, top-down approach grounded in energy-based interpretation. Specifically, we formalize token dynamics as a joint maximum likelihood estimation on the hypersphere, featuring two properties: semantic alignment in the high-dimensional space and distributional uniformity in the low-dimensional space. By quantifying them with extended Hopfield energy functions, we instantiate this idea as a constrained energy minimization problem, which enables designs of symmetric attention and feedforward modules with RMS normalization. We further present \textit{Hyper-Spherical Energy Transformer} (Hyper-SET), a recurrent-depth alternative to vanilla Transformers naturally emerging from iterative energy optimization on the hypersphere. With shared parameters across layers, Hyper-SET can scale to arbitrary depth with fewer parameters. Theoretically grounded and compact, it achieves competitive or superior performance across diverse tasks, including Sudoku solving, image classification, and masked image modeling. We also design novel variations under the proposed general principle, such as linear attention and gated feedforward layer. Moreover, we showcase its scalability with depth-wise LoRA. Our results highlight Hyper-SET as a step toward interpretable and principled Transformer design.

\end{abstract}

\section{Introduction}
\label{section:intro}
Transformer-based models \cite{NIPS2017_3f5ee243} have become foundational across diverse domains, including computer vision \cite{dosovitskiy2021an,bao2022beit,he2022masked,peebles2023scalable}, natural language \cite{devlin-etal-2019-bert,Lan2020ALBERT, brown2020language}, robotics \cite{brohan2022rt,chen2021decision}, and scientific discovery \cite{jumper2021highly, kamienny2022end}. In recent years, there has been evidence that scaling up model size, dataset size, or computational budget during pre-training can yield unprecedented performance gains \cite{kaplan2020scalinglawsneurallanguage}, driving the proliferation of Transformer-based foundation models \cite{openai2024gpt4technicalreport,dubey2024llama,anil2023gemini,oquab2024dinov}. 

Despite these achievements, the architecture of Transformers—especially the configuration and role of individual layers—remains largely heuristic. For instance, empirical studies have observed high redundancy in the deeper layers \cite{gromov2024unreasonable,men2024shortgpt}, uniformity of representations in the middle layers \cite{sun2024transformer}, and robustness to permuting certain intermediate layers \cite{lad2024remarkable} in LLMs. These findings suggest convergent functionality that one layer represents, yet our understanding of its role in processing information and representation learning remains limited. While interpretability efforts to unveil the function underlying the network layers exist, especially Transformer blocks-ranging from mechanistic interpretability \cite{elhage2021mathematical,nanda2023progress,wang2023interpretability,huben2024sparse} to causal mediation analysis \cite{vig2020investigating,meng2022locating} and visualization \cite{bricken2023monosemanticity,olsson2022context}-most focus on \textit{post hoc} interpretation and phenomenological approaches. This motivates a pivotal question: 
% \textit{Can we find or design a function prior that induces a model that is interpretable by construction?}
\begin{tcolorbox}
[highlightquestion, boxrule=1pt]
    \centering
    \textit{Can we find or design a function prior that induces a model that is interpretable by construction?}
\end{tcolorbox}

One approach to achieving intrinsic interpretability is to embed an explicit optimization process into neural networks, known as model-based deep learning \cite{shlezinger2023model}. Prior works have designed networks that solve domain-specific problems such as constraint satisfaction \cite{wang2019satnet}, optimal control \cite{amos2017optnet, amos2018differentiable}, or physical simulation \cite{greydanus2019hamiltonian,karniadakis2021physics}. However, these models often rely on fixed task priors and lack generality.

Another more general avenue is energy-based learning (EBL) \cite{dawid2024introduction}, which frames prediction as minimizing a scalar energy function $E_\theta(x, y)$ over outputs $y$ conditioned on inputs $x$. Within this framework, Energy Transformer \cite{hoover2024energy} interprets Transformer layers as iterative optimization over the canonical continuous Hopfield energy \cite{ramsauer2021hopfield,krotov2021large} yet focuses on mechanistic analogies to associative memory without grounding its formulation in specific representational challenges.
In contrast, our goal is to find a design principle from top down that can not only reinterpret existing components but is also generalizable to novel architecture design \textit{constructively}.

In this work, we therefore take a fundamentally different approach by introducing a principle grounded in maximum likelihood estimation (MLE) for tokens on the hypersphere. Under mild assumptions, we interpret it under two complementary objectives for representation dynamics: semantic alignment (mode seeking) in high-dimensional space and distributional uniformity (mass covering) in a low-dimensional subspace. To translate these objectives into optimizable quantities over tokens, we define two complementary Hopfield-style energy functions that quantify these objectives and can be minimized through iterative optimization. This leads to the \textit{Hyper-Spherical Energy Transformer} (Hyper-SET)—a recurrent-depth model in which core components such as symmetric attention, feedforward layers, $\operatorname{RMSNorm}$, and skip connection emerge naturally from the optimization dynamics. With only one set of shared parameters across iterations, Hyper-SET is compact, interpretable by design, and empirically competitive across diverse tasks, including reasoning, classification, and masked image modeling. Beyond a single instantiation, this principle can induce novel architectural designs by generalizing the energy functions, enabling variants such as linear attention and gated feedforward layer. To enhance scalability, we introduce depth-wise low-rank adaptation (LoRA), allowing flexible iteration-specific modulation with minimal parameter overhead. Our key contributions are summarized as follows:
\begin{enumerate}[leftmargin=*]
\item \textbf{Theoretical Formulation}: We conceptualize a general principle based on maximum likelihood estimation on the hypersphere, quantified via complementary Hopfield-style energy functions.
\item \textbf{Energy-Driven Architecture}: We derive a compact Transformer-based model through pure energy minimization, where core components—including symmetric attention, feedforward, RMSNorm~\cite{zhang2019root}, and skip connection—emerge naturally.
\item \textbf{Competitive Performance}: We show competitive performance to vanilla Transformer across reasoning, classification, and masked modeling while demonstrating generality to design novel components (\eg, linear attention, gated feedforward) and scalability with flexible computation.
\end{enumerate}
% \todo{Add that we can write down the explicit energy instead of those who define the energy implicitly.}

\section{Related Work}
\label{section:label}

\subsection{Energy-based Learning}
Energy-based learning (EBL) \cite{lecun2006tutorial,dawid2024introduction} provides a unifying framework for modeling prediction as minimizing an energy function. Early forms include Hopfield networks \cite{hopfield1982neural} and Boltzmann machines \cite{ackley1985learning}. Modern developments in EBL span both generative modeling—via energy functions \cite{du2019implicit} or their gradients (as in score-based models \cite{pmlr-v37-sohl-dickstein15,song2019generative})—and representation learning. Another line of work views network layers as the result of iterative energy minimization. Some approaches define energy implicitly through neural networks \cite{bai2019deep,du2022learning, Du_2024_ICML}, while recent work Energy Transformer \cite{hoover2024energy} draws analogies between attention layers and explicit energy descent but mainly focuses on reinterpretation rather than principled derivation. Our work differs in that we design the Transformer block by quantifying a general principle that can induce variants through alternative energy.

Other works also explore energy formulations on the hypersphere \cite{liu2018learning,loshchilov2024ngpt}, but mostly in the weight space. By contrast, we define our energies directly on the representation space. Additionally, recent theoretical studies on memory capacity in modern Hopfield networks \cite{hu2024computational,pmlr-v235-wu24i,hu2024provably} emphasize spreading patterns on the sphere but focus primarily on memory retrieval and cross-attention. 

% aligning with recent theoretical work on memory retrieval in Hopfield networks \cite{hu2024computational,pmlr-v235-wu24i}. However, those focus on capacity analysis in memory retrieval settings and are more akin to cross-attention. Our formulation targets representation learning and token dynamics in self-attention settings.

\subsection{Model Design from First Principles}
While neural network architectures are often shaped by engineering practices, recent work has explored designing or interpreting them through principled lenses like signal processing, information theory, and neurobiology. For example, deep unrolling of the sparse coding algorithms has led to the development of fully connected networks \cite{gregor2010learning}, convolution networks \cite{papyan2017convolutional, papyan2018theoretical}, and even graph neural networks through iterative algorithms \cite{yang2021graph}. Similarly, the sparse rate reduction principle has been used to derive the Transformer architecture \cite{yu2023white}. Other approaches draw inspiration from approximation theory \cite{liu2024kan} and brain computation \cite{kozachkov2023building}, further bridging the gap between theoretical insights and practical network design. 
% Add later if possible
% Our work takes on a dynamical system view to design more interpretable Transformers in the sense that the forward dynamics can be derived from maximum likelihood and described by meaningful energy functions.

% \subsection{Theoretical Understanding of Transformer}
% \subsection{Recurrence in Transformer}
% universal transformer, albert, R-transformer, ...
% There are also arguments for improving compositional generalization with universal transformers (weight-tying like ours) \cite{ontanon-etal-2022-making,petty2024impact}.

\begin{figure}[tbp]
% \vskip -.1in
    \centering
    \includegraphics[width=\linewidth]{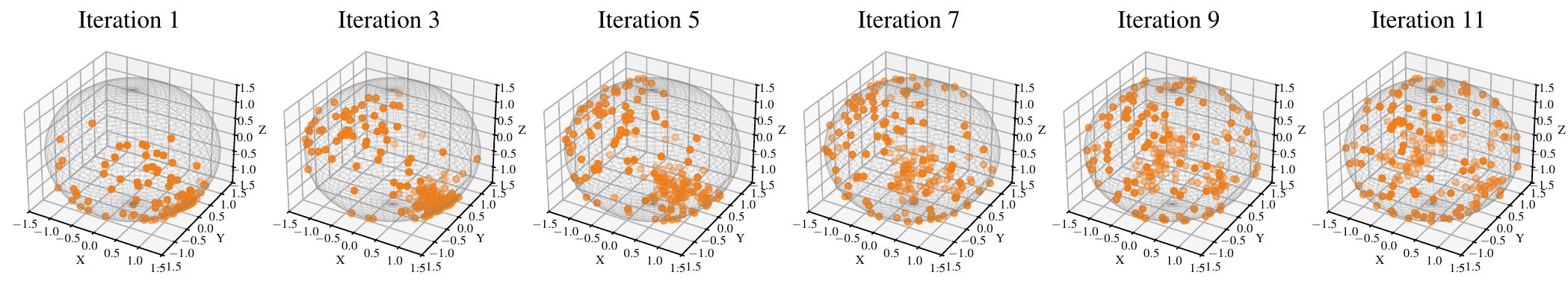}
    \includegraphics[width=\linewidth]{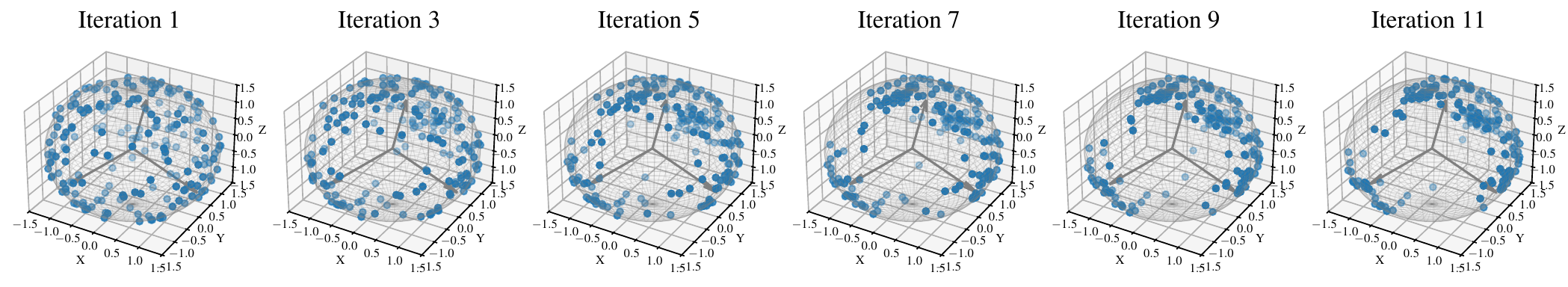}    
        % \vskip -.2in
    \caption{Evolution of tokens in the forward pass. \textit{Top}: Tokens projected onto subspaces are progressively separated on the \textbf{low-dimensional} hypersphere. \textit{Bottom}: Tokens gradually align with anchor vectors in the \textbf{high-dimensional} hypersphere. Visualization is carried out in three-dimensional space for illustrative purposes.}
    \label{fig:dynamics_on_hypersphere}
    % \vskip -.2in
\end{figure}
\section{Conceptualization and Instantiation}
% \todo{build connections to contrastive learning; should cite figure 1 here instead of in section 4}
% \todo{mixture of vMF distribution, which is defined on a sphere}
% Understanding Contrastive Representation Learning through Alignment and Uniformity on the Hypersphere
% A Cookbook of Self-Supervised Learning

% \subsection{Conceptualization}
% The objective~\eqref{eq:mle ultimate form} suggests that for a representation vector $\xvector \in \mathbb{R}^d$, the forward dynamics of a neural network can be characterized by two complementary properties: 1) seeking alignment with a set of directions in the \textbf{high-dimensional space} that encapsulate specific information derived from data per se, i.e. \textbf{mode seeking}; 2) maximally preserving the entropy embedded in the \textbf{low-dimensional space} via regularizing distribution uniformity, i.e., \textbf{mass covering}. 
To answer the introductory question, we conjecture that effective representations should exhibit two complementary properties: \textbf{semantic alignment} in a high-dimensional space and \textbf{distributional uniformity} in a low-dimensional subspace. This dual perspective reflects the balance of \textit{mode seeking} and \textit{mass covering}—terms we use to characterize the interplay between information preservation and entropy collapse prevention in representation learning.
An illustrative example is shown in Figure~\ref{fig:dynamics_on_hypersphere}.
% \subsection{Instantiation}

We formalize this conceptualization under maximum likelihood estimation. Specifically, we instantiate the forward dynamics as an optimization over a token-level vector $\xvector$ balancing two terms:
\begin{equation}
\label{eq:mle ultimate form}
\min_{\xvector} \quad \sum_{h=1}^H \underbrace{\operatorname{D_{KL}}( p(\zvector) \parallel p_\phi(\zvector^h | \xvector) )}_{\text{uniformity}} \underbrace{- \log p_\theta(\xvector)}_{\text{alignment}},
\end{equation}
where $\zvector^h$ represents low-dimensional projections of the high-dimensional representation $\xvector$. 

The first term encourages the projections $\zvector^h$ to approximate a prior uniform distribution $p(\zvector)$ on a hypersphere, thus maximizing entropy and mitigating representational collapse. The second term promotes alignment between $\xvector$ and mean directions, which can be modeled using von Mises–Fisher distributions. A detailed justification and interpretation of this objective is provided in Appendix~\ref{section:justification}.

This objective resonates with but differs from the contrastive learning objective that unifies alignment and uniformity in a shared latent space \cite{wang2020understanding}. Our work instead takes on an energy view to quantify these two key ingredients into optimizable functions of $\xvector$ that can induce Transformer architectures.

\section{Hyperspherical Energy Transformer from Iterative Energy Minimization}
\begin{figure*}[tbp]
    \centering
    \includegraphics[width=\textwidth]{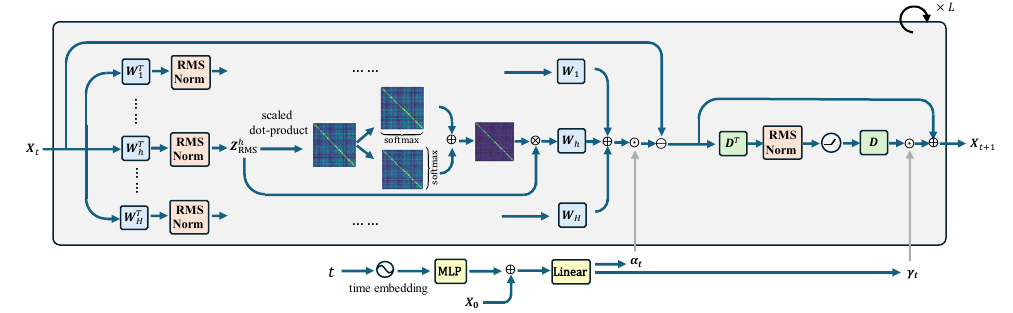}
    % \vskip -0.2in
    \caption{Overview of our hyperspherical energy Transformer layer. It recovers sequential stacking of symmetric self-attention, feedforward, skip connection, and $\operatorname{RMSNorm}$ from sheer minimization of extended Hopfield energy. Adaptive step sizes are learned given the current $t$ and initial input $\Xmatrix_0$.}
    % \vskip -.2in
    \label{fig:overview}
\end{figure*}

% \todo{We take an energy perspective on mle and manifest how to translate this formulation into Transformer architecture by leveraging Hopfield function to quantize these two terms}
% \todo{how to bring in the definition of Emch}
In this section, we translate the proposed instantiation into two modified Hopfield energy functions defined on hyperspheres (see Appendix~\ref{section:preliminaries} for preliminaries and definition of Hopfield energy $E_\text{MCH}$ in~\eqref{eq:mch}). Through iterative energy minimization, the architectural components of Transformer layers naturally arise under this framework. The overview is presented in Figure~\ref{fig:overview}.   

% \vspace{-0.5cm}
\subsection{Hyperspherical Energy}
\label{section:hyperspherical energy}
% We consider $N$ vectors $\Xmatrix = [\xvector_1, \dots, \xvector_N]$ from a probabilistic space with $\xvector_i \in \mathbb{R}^d, i\in [N]$, which can be seen as the contextual tokens in Transformers, and denote two different sets of basis vectors of this $d$-dimensional space, $\Wmatrix = [\Wmatrix_1, \dots, \Wmatrix_H] \in \mathbb{R}^{d\times Hp}$ and $\Dmatrix = [\dvector_1, \dots, \dvector_M] \in \mathbb{R}^{d\times M}$. Here, $\Wmatrix_i \in \mathbb{R}^{d\times p}$ represents the projection to the $p$-dimensional subspace. Unless otherwise specified, we assume the column vectors are incoherent and span the full space, i.e., $Hp=M=d$.
Let $\Xmatrix = [\xvector_1, \dots, \xvector_N]$ denote a set of $N$ contextual token vectors, each $\xvector_i \in \mathbb{R}^d$. These tokens are projected into $H$ distinct subspaces via basis matrices $\Wmatrix = [\Wmatrix_1, \dots, \Wmatrix_H] \in \mathbb{R}^{d \times Hp}$, where each $\Wmatrix_h \in \mathbb{R}^{d \times p}$ spans a $p$-dimensional subspace. Additionally, we define a second set of bases $\Dmatrix = [\dvector_1, \dots, \dvector_M] \in \mathbb{R}^{d \times M}$ to encode semantic directions in the original space. Unless otherwise specified, we assume these basis vectors are incoherent and span the full space, \ie, $Hp = M = d$.

\subsubsection{Overcoming Token Synchronization via Repulsive Dynamics}
% A recent study \cite{yu2023white} argues that the contextual tokens lie on a low-dimensional manifold of their high-dimensional ambient space. We adopt this view and study the projection of tokens with bases $\Wmatrix$; for a subspace spanned $\Wmatrix_h$, the latent representation of token $\xvector_i$ can be written as: 
Motivated by a recent argument that the contextual tokens lie on a low-dimensional manifold of their high-dimensional ambient space \cite{yu2023white}, we study the projection of tokens with bases $\Wmatrix$; for a subspace spanned by $\Wmatrix_h$, the latent representation of a token $\xvector_i$ can be written as 
\begin{equation}
\label{eq:1}
    \zvector_i^h = \Wmatrix_h^T\xvector_i.
\end{equation}
% \todo{refer to appendix for definition}
Canonical Hopfield energy $E_\text{MCH}$ tends to align vectors with stored patterns. This interaction often occurs between dynamic tokens and static patterns. However, in Transformers' self-attention, this interplay happens among all dynamic tokens simultaneously. Enforcing strict alignment among them risks collapsing representations into degenerate clusters, reducing expressiveness. This phenomenon has been observed empirically as oversmoothing \cite{chen2022principle,wu2024demystifying} or rank collapse \cite{pmlr-v139-dong21a}, and theoretically characterized in \cite{geshkovski2023the}. It also relates to the synchronization effect in coupled systems \cite{acebron2005kuramoto,miyato2024artificial}. 

Therefore, to overcome this issue, we extend the Hopfield energy $E_\text{MCH}$ to model the repulsive force among tokens and quantify their distributional uniformity in each subspace, which serves as a surrogate of the uniformity measure in~\eqref{eq:mle ultimate form}. For subspace $h$, this energy is given by
\begin{equation}
\label{eq:2}
    E_{\text{ATTN}}^h = \beta^{-1}\sum_{i=1}^N\log\left(\sum_{j=1}^N \exp\left(\beta(\zvector_i^h)^T(\zvector_j^h)\right)\right),
\end{equation}
where $\beta$ is usually the inverse of temperature. Here we use the subscript $_\mathrm{ATTN}$ as this energy will be shown to be related to the design of the attention layer, resembling that in \citep{yu2023white}. Aggregating over all subspaces, the total energy that models interacting tokens would be 
\begin{equation}
\label{eq:3}
        E_{\text{ATTN}} = \sum_{h=1}^H E_{\text{ATTN}}^h, \quad\text{subject to} ~\|\Wmatrix_h^T\xvector_i\|_2 = \sqrt{p}. 
\end{equation}
The constraint ensures that the dynamics take place on a low-dimensional hypersphere of radius $\sqrt{p}$. Minimizing $E_{\text{ATTN}}$ thus encourages token spread evenly on multiple hyperspheres, mitigating collapse and promoting distributional uniformity. \footnote{Its asymptotic convergence to uniformity on spheres has been proved in \cite{liu2018learning}.}
% \footnote{Its asymptotic convergence to uniformity on spheres has been proved \cite{liu2018learning}.}
% \footnote{The connection of its optimality to uniformity on spheres has been studied as the Thomson problem \cite{liu2018learning}.}

% By minimizing~\eqref{eq:3}, the tokens are encouraged to be distributed on the sphere as uniformly as possible. 
% \todo{any derivations or reference to connect its optimality to uniformity of sphere;}
% An illustrative example is shown in Figure~\ref{fig:dynamics_on_hypersphere}.

\subsubsection{Semantic Alignment via Attraction to High-Dimensional Bases}
% The tokens projected to subspaces separate to occupy more volume and enrich the information they encode. 
While the subspace projections separate to occupy more volume thus regularizing distribution, we seek to enrich the high-dimensional representations per se. From an information-theoretic perspective \cite{tishby2000information, tishby2015deep}, effective representations require compressing uninformative redundancy while preserving salient information. Hence, in the original high-dimensional space, we encourage token alignment with a set of directions that contain knowledge from data to reduce entropy for minimal coding bits.

% This implies that tokens in the original space should coalesce into several distinct clusters. 
Motivated by empirical findings that the feedforward layers in Transformers store much of the model’s knowledge \cite{geva-etal-2021-transformer, dar-etal-2023-analyzing}, we interpret the basis vectors $\Dmatrix$ as the semantic directions. One surrogate function to implement this attractive energy for alignment in~\eqref{eq:mle ultimate form} is defined as 
\begin{equation}
\label{eq:4}
    E_{\text{FF}} =  -\frac{1}{2}\sum_{i=1}^N\sum_{m=1}^M\left(\operatorname{ReLU}\left(\dvector_m^T\xvector_i\right)\right)^2, \quad \text{subject to} ~\|\Dmatrix^T\xvector_i\|_2 = \sqrt{M}.
\end{equation}
Here we use the subscript $_\mathrm{FF}$ as this energy relates to the design of the feedforward layer. This energy favors alignment between tokens and those basis directions forming acute angles (as filtered by ReLU), while maintaining the hyperspherical constraint in the original space. Geometrically, each token is drawn toward a union of attractive half-spaces defined by $\Dmatrix$.
% By minimizing this energy, a token tends to lie on the union of half-spaces defined by basis vectors that form an acute angle with this token while still residing on the hypersphere of the original space. 

\subsubsection{Dual Energy on the Hypersphere}
\label{section:symmetric structure induced from energy minimization}
By combining these two hyperspherical energy functions, we introduce a unified objective function that characterizes the functionality the Transformer layer represents:
\begin{align}
\label{eq:5}
    \min_{\xvector_1, \dots,\xvector_N \in \Xmatrix} \quad & E(\Xmatrix ; \Wmatrix, \Dmatrix) = E_{\text{ATTN}} + E_{\text{FF}} \\
    \text{subject to} \quad & \|\Wmatrix_h^T\xvector_i\|_2 = \sqrt{p}, \quad \|\Dmatrix^T\xvector_i\|_2 = \sqrt{M}, \quad i = 1, \dots, N. \nonumber
\end{align}
Iteratively minimizing this energy under spherical constraints induces the core architecture of Transformer layers: self-attention module arises from repulsive energy over subspaces and feedforward module arises from attractive energy in the ambient space. To solve optimization~\eqref{eq:5}, we adopt an alternating minimization method by splitting it into sub-problems, following \cite{yu2023white}. 
% This characterization of Transformer optimization problem somewhat resonates with the \textit{compression-sparsification} procedure in \cite{yu2023white}, where they frame the objective as compressing information of tokens in the subspaces and enlarging volume in the original space. Yet our objective can be interpreted from the opposite direction. We aim to gradually enrich the information in the subspaces while peeling off redundancy in the original space. This also mirrors the rationale of minimal and sufficient statistics from the information bottleneck \cite{tishby2015deep}. To solve optimization~\eqref{eq:5}, we consider an alternating minimization method by splitting it into sub-problems, following \cite{yu2023white}. 

\subsection{Symmetric Structure Induced From Energy Minimization}
\subsubsection{Attention Module from Uniform Energy}
\label{section:attention}
To show how we have an attention module derived from minimizing hyperspherical energy $E_{\text{ATTN}}$ in~\eqref{eq:3}, we first establish the differential equation that models the evolution of tokens' interactions: 
\begin{equation}
\label{eq:6}
\begin{split}
\dot{\Xmatrix} &= -\nabla_{\Xmatrix} E_{\text{ATTN}}   \\
&=-\sum_{h=1}^H\Wmatrix_h\Wmatrix_h^T\Xmatrix\left( \underbrace{\operatorname{softmax}}_{\text{column-wise}}\left(\beta (\Wmatrix_h^T\Xmatrix)^T(\Wmatrix_h^T\Xmatrix)\right)+ \underbrace{\operatorname{softmax}}_{\text{row-wise}}\left(\beta (\Wmatrix_h^T\Xmatrix)^T(\Wmatrix_h^T\Xmatrix)\right) \right)
\end{split}   
\end{equation}
where $\beta=1/\sqrt{p}$ as in vanilla Transformers \cite{NIPS2017_3f5ee243}. Derivations could be found in Appendix~\ref{section:derivation of E attn}. 

The constraint on the low-dimensional hypersphere of radius $\sqrt{p}$ corresponds to $\operatorname{RMSNorm(\cdot)}$, which bears resemblance to Query-Key Normalization \cite{henry-etal-2020-query}, but here the normalization is applied after projection by the same query-key-value matrix. The projections in subspace $h$ onto the hypersphere thus read as 
\begin{equation}
\label{eq:7}
\Zmatrix_{\text{RMS}}^h = \operatorname{RMSNorm}(\Zmatrix^h) = \operatorname{RMSNorm}(\Wmatrix_h^T\Xmatrix).
\end{equation}
By discretizing the differential equation~\eqref{eq:6} with step size $\alpha_t$ and maintaining the constraint~\eqref{eq:7}, 
we obtain an self-attention module; let $[\Qmatrix\Kmatrix]_{\text{RMS,t}}= \beta (\Zmatrix_{\text{RMS},t}^h)^T (\Zmatrix_{\text{RMS},t}^h) $, then the update will be:
% \begin{equation}
% \label{eq:8}
% \begin{split}
% \Xmatrix_{t+1} 
% &=\Xmatrix_{t} - \alpha_t \sum_{h=1}^H\left( (\Wmatrix_h\Zmatrix_{\text{RMS},t}^h \underbrace{\operatorname{softmax}}_{\text{column-wise}}\left(\beta (\Zmatrix_{\text{RMS},t}^h)^T(\Zmatrix_{\text{RMS},t}^h)\right) + \right.\\
% & \left. \Wmatrix_h\Zmatrix_{\text{RMS},t}^h\underbrace{\operatorname{softmax}}_{\text{row-wise}}\left(\beta (\Zmatrix_{\text{RMS},t}^h)^T(\Zmatrix_{\text{RMS},t}^h)\right) \right) 
% \end{split}   
% \end{equation}
\begin{equation}
\label{eq:8}
\Xmatrix_{t+1} =\Xmatrix_{t} - \alpha_t \sum_{h=1}^H \Bigg( \Wmatrix_h\Zmatrix_{\text{RMS},t}^h \underbrace{\operatorname{softmax}}_{\text{column-wise}} \left([\Qmatrix\Kmatrix]_{\text{RMS,t}}\right) + \Wmatrix_h\Zmatrix_{\text{RMS},t}^h \underbrace{\operatorname{softmax}}_{\text{row-wise}} \left( [\Qmatrix\Kmatrix]_{\text{RMS,t}} \right) \Bigg).
\end{equation}
% This update realizes a doubly symmetric attention operator with a multi-head structure from $H$ subspace partition. It features a highly symmetric structure in the sense that both the query-key dot product and attention weights are symmetric under row and column operations. Notably, this formulation connects to recent results on Wasserstein gradient flows using doubly stochastic attention~\cite{pmlr-v151-sander22a}, further grounding attention in energy-based principles. Here we offer another variant of attention that meets the symmetric query-key dot-production assumption therein. 

This update yields a doubly symmetric multi-head attention operator, where both the query-key dot product and attention weights are symmetric under row and column operations. This structure connects with formulations of Wasserstein gradient flows using doubly stochastic attention~\cite{pmlr-v151-sander22a}, grounding our energy-based interpretation. 

% We also present a variant compatible with the symmetric query-key assumption therein.

% This update brings up a new attention module with skip connection, and it has highly symmetric structures. On one hand, the query-key dot product is symmetric, while at the same time, the attention matrix is symmetric as well due to the sum of $\operatorname{softmax}$ in both column and row directions. The sum over $H$ different subspaces can be understood as the multi-head structure. 

% Another interesting connection with prior work is that a doubly stochastic attention matrix has proven to have equivalence to the Wasserstein gradient flow of some global energy \cite{pmlr-v151-sander22a}. Here we offer another variant of attention that meets the symmetric query-key dot-production assumption therein and can also be seen as the discretization of an explicit energy function.
\subsubsection{Feedforward Module from Alignment Energy}
\label{section:section ff}
For the sub-problem of minimizing the alignment energy $E_{\text{FF}}$ in~\eqref{eq:4}, we have a similar construction of the corresponding differential equation, with details deferred to Appendix~\ref{section:derivation of E ff}:
\begin{equation}
\label{eq:9}
\dot{\Xmatrix} = -\nabla_{\Xmatrix} E_{\text{FF}} = \Dmatrix\operatorname{ReLU}\left(\Dmatrix^T\Xmatrix\right).
\end{equation}

By further imposing the high-dimensional hyperspherical constraint via $\operatorname{RMSNorm}$ with discretization step size $\gamma_t$, we can recover the feedforward layer that exhibits symmetry in the weight space:
\begin{equation}
\label{eq:10}
\Xmatrix_{t+1} = \Xmatrix_t + \gamma_t \Dmatrix\operatorname{ReLU}\left(\operatorname{RMSNorm}\left(\Dmatrix^T\Xmatrix \right)\right)  
\end{equation}

% Notice that this feedforward module also bears symmetry in the weight space. 
\subsection{Learning Adaptive Step Size}
To make the step sizes more flexible, we choose to learn their embedding with a neural network conditioned on the current iteration $t$ and the initial token $\xvector(0)$ (usually the output of the tokenizer):
\begin{equation}
\alpha_t = \bm{\alpha}_\eta (t, \xvector(0)),   
\quad\gamma_t = \bm{\gamma}_\psi (t, \xvector(0)).     \label{eq:12}
\end{equation}

For each iteration, step size embeddings in~\eqref{eq:12} are applied channel-wise to each token, similar to techniques in \cite{touvron2021going,peebles2023scalable} and detailed in Appendix~\ref{section:Network to Learn Adaptive Step Size}. We also adopt the zero-initialization of network parameters $\eta$ and $\psi$ from \cite{bachlechner2021rezero} to facilitate convergence when using larger iterations. 

In summary, by combining all the components and techniques, we present the \textit{Hyper-Spherical Energy Transformer} (Hyper-SET) with attention and feedforward sequentially stacked and with only one layer of learnable parameters. This one-layer model is amenable to rigorous analysis and, as we will demonstrate later, has competitive performance with vanilla Transformer.

\section{Experiment}
% \subsection{Synthetic Data}
% \subsubsection{Matrix Completion}
In this section, we evaluate Hyper-SET against vanilla Transformer and other baselines on discriminative and generative tasks. For fairness, we remove biases and dropout, use the Pre-Norm style with $\operatorname{RMSNorm}$, and set the MLP ratio to 4 in Transformer. We use one-layer trainable parameters but vary the forward iterations for all models, including Transformers, unless otherwise specified.
% As one iteration of energy minimization corresponds to a single-layer update of tokens in Hyper-SET, we use one-layer trainable parameters but vary the iteration for all models, including Transformers, unless otherwise specified.
\footnote{For instance, 12 iterations mean applying the layer repeatedly 12 times.}
% All experiments fit the memory size of an 80GB NVIDIA A100 GPU. 

\begin{figure}[tbp]
  \centering
  \begin{subfigure}[t]{0.49\textwidth}
    \centering
    \includegraphics[width=\linewidth]{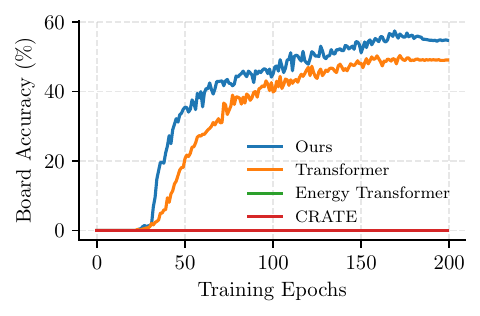}
    \caption{Sudoku training dynamics}
    \label{fig:sudoku training}
  \end{subfigure}
  \hfill % Adds horizontal space between subfigures
  \begin{subfigure}[t]{0.49\textwidth}
    \centering
    \includegraphics[width=\linewidth]{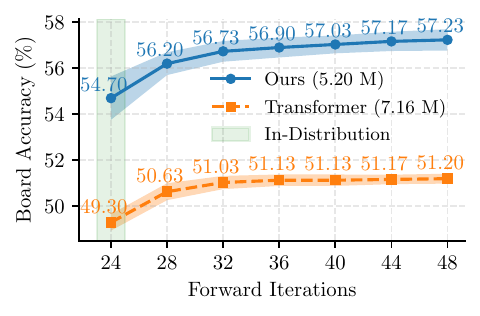}
    \caption{Sudoku test-time extrapolation}
    \label{fig:test-time extrapolation}
  \end{subfigure}
  \caption{\textit{Left}: Training dynamics of different approaches. Ours has a superior training curve and converges faster, while Energy Transformer and CRATE both fail to make accurate predictions. \textit{Right}: Test-time extrapolation w.r.t. the number of forward iterations. Our model achieves better performance, averaged over five runs, with fewer parameters even when the iterations are beyond the training regime. }
  \label{fig:sudoku training dynamics test-time extrapolation}
\end{figure}
\subsection{Solving Sudoku} 

% \begin{figure}[htbp]
%     \centering
%     \begin{minipage}[t]{0.49\textwidth}
%         \centering
%         \includegraphics[width=\linewidth]{sudoku_extropolation.pdf}
%         \caption{Sudoku training dynamics.}
%         \label{fig:sudoku training}
%     \end{minipage}
%     \hfill
%     \begin{minipage}[t]{0.49\textwidth}
%         \centering
%         \includegraphics[width=\linewidth]{sudoku_extropolation.pdf}
%         \caption{Test-time Extrapolation on Sudoku Board Accuracy w.r.t the Number of Forward Iterations. Our model achieves better performance with fewer parameters, even when the iterations are beyond the training regime. Results are averaged over five runs. }
%         \label{fig:test-time extrapolation}
%     \end{minipage}
% \end{figure}

\paragraph{Setup} We use the challenging dataset from \cite{palm2018recurrent}, featuring boards with only 17 to 34 known digits. We build on the code \footnote{\href{https://github.com/azreasoners/recurrent_transformer}{https://github.com/azreasoners/recurrent\_transformer}} from \cite{yang2023learning} and follow the setting of training on 9k samples and evaluating on 1k. Cross-entropy loss is computed exclusively on unknown entries. We train all models with 200 epochs, 16 batch size, AdamW \cite{loshchilov2018decoupled} with 0.1 weight decay, and learning rate from 1e-4 with cosine decay. The hidden dimension is set to 768. See Appendix~\ref{section:sudoku setup} for details. 
% \begin{figure}[t]
%     \centering
%     \includegraphics[width=\linewidth]{energy_sudoku.pdf}
%     \includegraphics[width=\linewidth]{energy.pdf}
%         % \vskip -.2in
%     \caption{The energy of both the attention and feedforward module decreases on Sudoku \textbf{(Top)} and CIFAR-10 \textbf{(Bottom)}, without hard constraints on the sign of the step size. This suggests the layer aligns well with the optimization objective. Normalization is first applied to meet the condition in \eqref{eq:5} before computing the energy.}
%     \label{fig:energy}
% \end{figure}

\paragraph{Extrapolation to Larger Iteration}
Under identical experimental conditions, our model exhibits faster and superior training dynamics over Transformer, while Energy Transformer \cite{hoover2024energy} and CRATE \cite{yu2023white} both fail on this task, as shown in Figure~\ref{fig:sudoku training}. It also outperforms Transformer for in-distribution evaluation (54.70$\%$ vs. 49.30$\%$), \ie, using the same forward iterations for training and inference. 

Recent efforts also explore test-time compute scaling to enhance reasoning \cite{schwarzschild2021can,bansal2022end,du2022learning,banino2021pondernet}, aiming to extrapolate the algorithms. Building on this idea, we increase test-time iterations up to 2× of training ones. As shown in Figure~\ref{fig:test-time extrapolation}, our model scales more effectively than Transformer, with larger accuracy gains. We attribute this extrapolation to learned adaptive step sizes that preserve energy minimization. In practice, we also find that trainable positional encoding is vital for the extrapolation. 

\begin{table}[htbp]
\caption{Top-1 accuracy on image classification. Models are under the same hidden dimension and 1 layer with 12 iterations. Scaling up our model to match Transformer's parameters gives better performance. Parameters are measured on CIFAR-10.}
\label{tab:classification}
\centering
% \resizebox{\linewidth}{!}{
\begin{tabular}{lllll}
\toprule
\multirow{2}{*}{Models}  & \multirow{2}{*}{Config ($\#$ Params)} & \multicolumn{3}{c}{Dataset}\\
\cmidrule(lr){3-5}
 &  & CIFAR-10 & CIFAR-100 & IM-100\\
\midrule
Transformer  & \texttt{Small} (1.79 M) & 89.90 & 61.89 &  69.44\\
CRATE-T \cite{hu2024an}  & \texttt{Small} (0.32 M) & 86.77 &  60.52 & 61.12 \\
CRATE \cite{yu2023white} & \texttt{Small} (0.47 M) & 86.97 & 61.13 & 63.52\\
Energy Transformer \cite{hoover2024energy} & \texttt{Small} (0.91 M) & 75.98 & 52.50 & 40.40\\
Ours  & \texttt{Small} (0.96 M) & 88.89 & 62.48 & 67.64\\
\midrule
% CRATE & Base (2.37466 M) & 84.94 & 53.75 & xxx  \\
Ours  & \texttt{Small} Scale-up (1.61 M) & \textbf{90.11} & \textbf{63.41} & \textbf{70.16}\\
\bottomrule
\end{tabular}
% }
\end{table}
\begin{figure}[tbp]
    \centering
    \includegraphics[width=\linewidth]{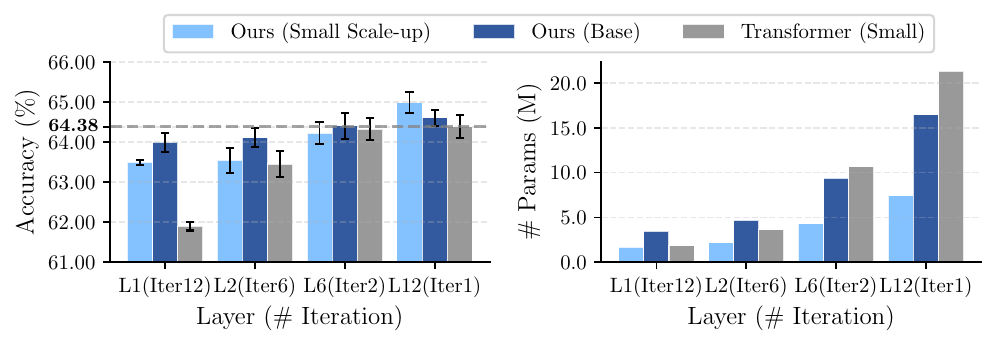}
    % \vskip -.2in
    \caption{Top-1 accuracy (\textit{Left}) and model parameters (\textit{Right}) on CIFAR-100 with different layer-iteration trade-offs. Error bars represent standard deviation over five runs.}
    \label{fig:bar plot cifar10}
\end{figure}
% \todo{need to adapt trade-off figure to the newest setting}

\subsection{Image Classification}
\label{section:image classification}

\paragraph{Setup \& Results}
We also evaluate Hyper-SET on CIFAR-10/100 and ImageNet-100 \footnote{We use a subset of ImageNet-1k from \href{https://github.com/HobbitLong/CMC/blob/master/imagenet100.txt}{\text{https://github.com/HobbitLong/CMC/blob/master/imagenet100.txt}}} comparing against ViTs, white-box Transformer CRATE \cite{yu2023white}, its variant CRATE-T that aims for more faithful implementations \cite{hu2024an}, and Energy Transformer \cite{hoover2024energy}. All models are trained with 200 epochs, 128 batch size, 12 forward iterations, Adam \cite{kingma2014adam} optimizer, cosine learning rate decay from 1e-3 with warm-up, and a learnable class token $[\texttt{CLS}]$. Detailed configurations are in Appendix~\ref{section:image classification setup}. 

Table~\ref{tab:classification} shows that, with the same hidden dimension ($\texttt{Small}$ configuration), our model surpasses others on CIFAR-100 but slightly lags behind Transformer on CIFAR-10 and ImageNet-100. Noticeably, our architecture reduces parameters by around 46$\%$ compared to the Transformer. When scaling up the hidden dimension to match the Transformer's parameters, our model achieves the best results.

\begin{table}[htbp]
\caption{Comparison of masked image modeling performance on ImageNet-100 (5k). Our model lags behind Transformer when given the same iterations, but matches its performance if scaling up them and the width of the feedforward module (larger $M$). Our model is also more parameter-efficient. }
\label{tab:masked image modeling}
\centering
\resizebox{\textwidth}{!}{
\begin{tabular}{llllcll}
\toprule
Models  &  Layer / Iteration / FF Ratio $M$ & PSNR ($\uparrow$) & SSIM ($\uparrow$) & Multi-Scale SSIM ($\uparrow$) & LPIPS ($\downarrow$)& FID ($\downarrow$)\\
\midrule
Transformer  & 1 / 12 / 4$d$ ~(8.85 M) & 15.953 & 0.417 & \textbf{0.599} & \textbf{0.327} & \textbf{43.428}
\\
Ours  & 1 / 12 / $d$ ~(3.94 M) & 15.713 & 0.411 & 0.576 & 0.358 & 59.841\\
Ours  & 1 / 24 / 8$d$ ~(8.07 M) & \textbf{15.955} & \textbf{0.417 }& 0.596 & 0.332 & 45.174\\
\bottomrule
\end{tabular}
}
\end{table}

\paragraph{Layer-Iteration Trade-off}
So far, the classification is conducted using a one-layer model. A natural question is how well the model performs when stacking multiple layers with different parameters. To see this, we first train a Transformer with 12 layers—equivalent in effective depth to one layer with 12 iterations—as an upper bound. We then vary the number of distinct layers and their iterations while keeping total depth constant, effectively introducing flexibility to the basis vectors. 

% Then, we vary the ratio of distinct layers and their respective iterations while maintaining the effective depth. This configuration can be interpreted as adding flexibility to the basis vectors. 
%which the energy is conditioned on. 

In Figure~\ref{fig:bar plot cifar10}, our scaled-up $\texttt{small}$ model has parameter efficiency across varied layer-iteration ratios, with this strength intensifying as more independent layers are trained. However, its architectural efficiency limits scalability beyond two layers. Scaling to the $\texttt{Base}$ configuration enables our model to consistently outperform Transformer, exceeding the upper bound while retaining parameter efficiency.

\subsection{Masked Image Modeling}
\paragraph{Setup \& Results}
Masked image modeling has recently regained its attention for autoregressive generation \cite{li2023mage,pmlr-v202-chang23b,li2024autoregressive}, framed as recovering images from $100\%$ masking. Due to its high computational demand, we attempt to demonstrate the power of our one-layer model specifically for image reconstruction on ImageNet-100. We build on prior work \cite{chang2022maskgit} and use the open-source repository. \footnote{\href{https://github.com/valeoai/Maskgit-pytorch}{\text{https://github.com/valeoai/Maskgit-pytorch}}} Concrete settings are in Appendix~\ref{section:masked image modeling setup}, with additional results and visualization in Appendix~\ref{section:additional results of masked image modeling}.

% \begin{wrapfigure}{r}{0.5\textwidth}
% \centering
% \includegraphics[width=\linewidth]{visualization.pdf}
% \caption{Visual Comparison on Masked Image Modeling on ImageNet 256$\times$256. Our model, when scaling to Transformer scale with additional compute, can achieve similar reconstruction quality when masking ratio $=40\%$.}
% \label{fig:visualization}
% \end{wrapfigure}

Table~\ref{tab:masked image modeling} unveils that, under the same number of iterations, our model significantly reduces parameters but lags behind Transformer on all metrics. If we further increase its iterations and the width of feedforward module $M$ to 8$d$, it can fill in the performance gap but at the cost of more computation.
% \begin{figure}[!tbp]
%     \centering
%     \includegraphics[width=\linewidth]{visualization.pdf}
%         % \vskip -.1in
%     \caption{Visual Comparison on Masked Image Modeling on ImageNet 256$\times$256. Our model, when scaling to Transformer scale with additional compute, can achieve similar reconstruction quality when masking ratio $=40\%$.}
%     \label{fig:visualization}
% \end{figure}

\begin{figure}[tbp]
  \centering
    \includegraphics[width=\linewidth]{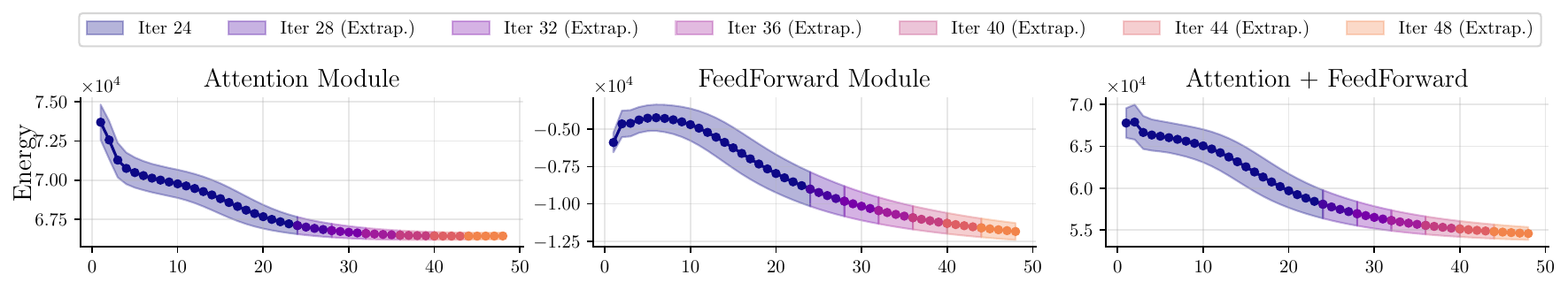}
    % \caption{Sudoku}
    % \label{fig:energy sudoku}
    % \vspace{-10pt}
    \includegraphics[width=\linewidth]{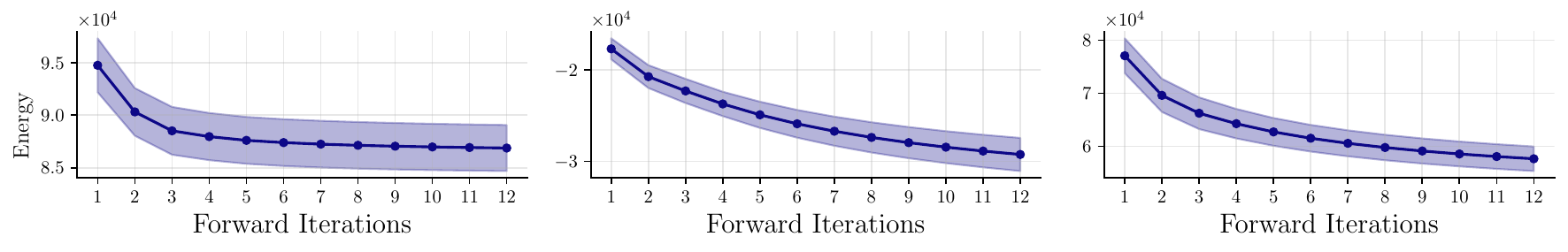}
    % \caption{CIFAR-10}
    % \label{fig:energy cifar10}
  \caption{The attention and feedforward energy decrease on Sudoku (\textit{Top}) and CIFAR-10 (\textit{Down}) even without sign constraints on the step sizes. This suggests the layer aligns well with the optimization objective. Normalization is first applied to meet the condition in \eqref{eq:5} before computing the energy.}
  \label{fig:energy}
\end{figure}
% \begin{figure}[tbp]
%   \centering
%   \begin{subfigure}[t]{0.49\textwidth}
%     \centering
%     \includegraphics[width=\linewidth]{energy_sudoku.pdf}
%     \caption{Sudoku}
%     \label{fig:energy sudoku}
%   \end{subfigure}
%   \hfill % Adds horizontal space between subfigures
%   \begin{subfigure}[t]{0.49\textwidth}
%     \centering
%     \includegraphics[width=\linewidth]{energy.pdf}
%     \caption{CIFAR-10}
%     \label{fig:energy cifar10}
%   \end{subfigure}
%   \caption{The energy of both the attention and feedforward module decreases on Sudoku (\textit{Left}) and CIFAR-10 (\textit{Right}), without hard constraints on the sign of the step size. This suggests the layer aligns well with the optimization objective. Normalization is first applied to meet the condition in \eqref{eq:5} before computing the energy.}
%   \label{fig:energy}
% \end{figure}

\begin{figure}[tbp]
  \centering
  \begin{subfigure}[t]{0.49\textwidth}
    \centering
    \includegraphics[width=\linewidth]{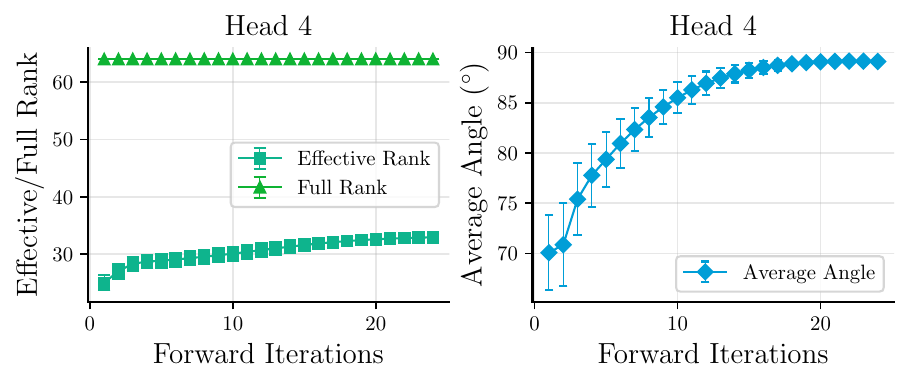}
    \caption{Sudoku}
    \label{fig:rank}
  \end{subfigure}
  \hfill % Adds horizontal space between subfigures
  \begin{subfigure}[t]{0.49\textwidth}
    \centering
    \includegraphics[width=\linewidth]{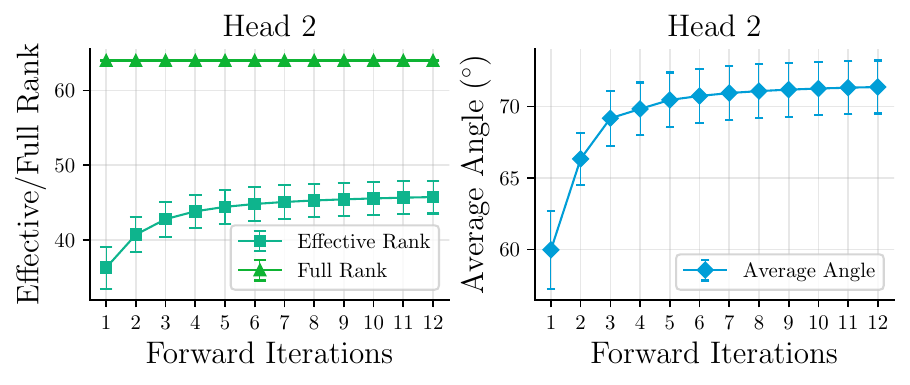}
    \caption{CIFAR-10}
    \label{fig:angle}
  \end{subfigure}
  \caption{The effective rank and average angle of tokens projected to one subspace gradually increase, suggesting a larger volume spanned by these tokens. Results are from Sudoku test dataset \cite{palm2018recurrent} (\textit{Left}) and CIFAR-10 validation set (\textit{Right}).}
  \label{fig:rank angle}
\end{figure}

% \begin{wrapfigure}{r}{0.5\textwidth}
% \centering
% \includegraphics[width=\linewidth]{energy_sudoku.pdf}
% \includegraphics[width=\linewidth]{energy.pdf}
% \caption{The energy of both the attention and feedforward module decreases on Sudoku (\textit{Top}) and CIFAR-10 (\textit{Bottom}), without hard constraints on the sign of the step size. This suggests the layer aligns well with the optimization objective. Normalization is first applied to meet the condition in \eqref{eq:5} before computing the energy.}
% \label{fig:energy}
% \end{wrapfigure}

\subsection{Energy Evolution, Effective Rank and Average Angle}
Figure~\ref{fig:energy} shows energy trajectories of the attention ($E_{\text{ATTN}}$) and feedforward module ($E_{\text{FF}}$). Even without a positive threshold for step sizes $\alpha_t$ and $\gamma_t$, the energy on Sudoku still decreases within training iterations and extrapolates smoothly beyond them, indicating strong generalization of learned step sizes. On CIFAR-10, our designed energy exhibits a monotonic decline for the $\texttt{Small}$ model too.

To verify our subspace uniformity objective, we track two metrics—\textit{effective rank} and \textit{average angle}—defined in Appendix~\ref{section:definition}. As shown in Figure~\ref{fig:rank angle}, the subspace effective rank steadily increases while the full rank remains unchanged. Meanwhile, average angles between tokens approach orthogonality, aligning with our goal to prevent entropy collapse. Full results and comparisons with parameter-sharing Transformer are in Appendix~\ref{section:rank angle sudoku cifar10 complete}.

\begin{table*}[htbp]
    \centering
    \begin{minipage}{0.55\textwidth}
    \centering
    \caption{Alternative designs on key components measured by top-1 accuracy on CIFAR-10 and CIFAR-100.}
    \label{tab:alternative design}
        \resizebox{\linewidth}{!}{
        \begin{tabular}{llcc}
        % \begin{tabularx}{\linewidth}{llll}
           \toprule
            \multirow{2}{*}{Components}  & \multirow{2}{*}{Alternative Designs} & \multicolumn{2}{c}{Dataset}\\
            \cmidrule(lr){3-4}
             &  & CIFAR-10 & CIFAR-100 \\
            \midrule
            \multirow{3}{*}
            {$E_\text{ATTN}$} & Bi-Softmax Attention (Ours)& \textbf{90.11} & \textbf{63.41} \\
               & Sigmoid Attention & 85.93 & 59.72 \\
               & Linear Attention & 84.88 & 56.97 \\
            \midrule
            \multirow{3}{*}{$E_{\text{FF}}$} & ReLU FF (Ours) & \textbf{90.11} &\textbf{63.41} \\
               & Softmax FF & 88.20 & 62.44 \\
               & Gated FF  & 84.99 & 59.29 \\
            \midrule
            \multirow{3}{*}{Step Size} & Learning Step Size (Ours)& \textbf{90.11} & \textbf{63.41} \\
               & $\alpha_t=\gamma_t$ = 0.5 & 25.81 & 57.92 \\
               % & $\alpha_t=\gamma_t$ = 0.2 & xx & xx \\
               & $\alpha_t=\gamma_t$ = 0.1 & 81.45 & 58.29 \\
            \bottomrule
        % \end{tabularx}
        \end{tabular}
        }
    \end{minipage}%
    \hspace{\fill}
    \begin{minipage}{0.44\textwidth}
    \centering
    \caption{Imagenet-100 accuracy of Hyper-SET under different matrix rank within LoRA. Depth-wise LoRA introduces flexibility in the computation at each iteration.}
    \label{tab:layer-wise lora}
        \resizebox{\linewidth}{!}{
        \begin{tabular}{lcc}
        % \begin{tabularx}{\linewidth}{lll}
            \toprule
            Rank  & $\#$ Params & Accuracy (\%)\\
            \midrule
            Ours & 1.93 M & 70.16 \\
            + depth-wise LoRA (r=4) & 2.03 M & 70.36\\
            + depth-wise LoRA (r=8) & 2.13 M & 70.40\\
            + depth-wise LoRA (r=16) & 2.33 M & 70.56\\
            + depth-wise LoRA (r=32) & 2.72 M & \textbf{72.20}\\
            \bottomrule
        % \end{tabularx}
        \end{tabular}
        }
    \end{minipage}
\end{table*}
\subsection{Alternative Designs and Scalability}
% A key strength of our formulation~\eqref{eq:mle ultimate form} is its generality: it is not limited to a single instantiation, but enables a spectrum of variants through alternative energy functions. By generalizing the attention energy with kernel functions, we can induce novel attention structures and even linear attention. Similarly, a gating mechanism can also arise from generalizing feedforward energy. Table~\ref{tab:alternative design} shows their evaluation along with different fixed step sizes. More elaboration is provided in Appendix~\ref{section:alternative designs}. 

% To further enhance the scalability of Hyper-SET while preserving its compact parameterization, we introduce an extension that adds lightweight flexibility to the shared layer parameters, inspired by~\cite{bae2025relaxed}. We insert learnable low-rank adapters for each forward iteration to modulate the shared weights in a depth-specific manner during training. Figrue~\ref{tab:layer-wise lora} shows that this flexibility adds up to the performance. We provide setup and other scaling results in Appendix~\ref{section:scalability}. 
A key strength of our formulation~\eqref{eq:mle ultimate form} lies in its generality—it supports a broad spectrum of model variants through alternative energy functions. For example, replacing the attention energy with a kernel-based function yields novel attention mechanisms, including linear attention. Similarly, gating in feedforward layers naturally arises by generalizing the feedforward energy. Table~\ref{tab:alternative design} summarizes the performance of these variants and results with different fixed step sizes, with details in Appendix~\ref{section:alternative designs}.

To improve scalability with compact parameterizations, we introduce a lightweight extension inspired by~\cite{bae2025relaxed}, where learnable low-rank adapters are added at each forward iteration to modulate shared weights. This depth-wise adaptation, shown in Table~\ref{tab:layer-wise lora}, enhances performance without significantly increasing parameter count. Setup and additional scaling results are included in Appendix~\ref{section:scalability}.

\section{Conclusion}
We present \textbf{Hyper-SET}, a Transformer architecture designed via iterative optimization of hyperspherical energy functions, bridging energy-based learning and practical model design. By formulating dual energy on the hypersphere under a general principle derived from maximum likelihood, Hyper-SET pursues distributional uniformity in the low-dimensional subspaces while promoting directional alignment with bases in the original high-dimensional space, constructing core Transformer components with intrinsic interpretability. Empirically, Hyper-SET matches or surpasses vanilla Transformers across diverse tasks with fewer parameters. Beyond a single architecture, our framework enables flexible design choices and scalable variants. This work contributes towards principled, more describable, and economical Transformer designs that are both theoretically motivated and practically effective. 

% This work aims to advance principled Transformer design, offering new perspectives on building more efficient, describable yet equally powerful Transformers that scale more economically grounded in optimization dynamics.

% \todo{Our work provides new perspectives on building more efficient and describable yet equally powerful Transformers that scale more economically.}
% Further improvements sparse \cite{tan2023sparse} or mixture-of-experts \cite{csordas2024moeut} techniques.

\section*{Limitations and Future Directions}
While Hyper-SET offers a principled and empirically competitive formulation for Transformer design, it also comes with several limitations that highlight directions for future work:

\begin{itemize}[leftmargin=*]
    \item First, the choice of subspace in our conceptualization is less explored. The choice of uniform prior on the hypersphere in assumption~\ref{assumption:2} could be too strong in practice. Overly enforcing uniformity may be restrictive in some tasks. Moreover, the number of subspaces $H$ and its dimension $p$ are chosen heuristically. Their relationship, if any, with the real data distribution remains unclear.
    \item Second, the modulation network to learn step sizes introduces complexity. Although we use a modulation network to learn step sizes adaptively, tuning the configuration of this network still requires considerable effort. In addition, it introduces more computational complexity despite that the overall architecture is more parameter-efficient.
    \item Third, our experiments on scalability are still preliminary. We confirm competitive and superior performance on less than 20 million parameters and prove depth-wise LoRA scaling effectiveness. However, extensions to truly large-scale settings—\eg, billion-level—have yet to be demonstrated. 
\end{itemize} 

We provide several promising future directions here and  other discussions along with broader impact in Appendix~\ref{section:discussion and limitations}:
\begin{enumerate}[leftmargin=*]
    \item \textbf{Autoregressive Modeling}: Hyper-SET currently lacks a causal structure, limiting its use in autoregressive sequence modeling. Adapting this principled design to GPT-style models with causal masking is an important future step.
    \item \textbf{Flow Matching and ODE Connections}: The iterative updates in Hyper-SET resemble neural ODEs, suggesting potential connections to flow matching techniques that may unify Transformer-based models with generative modeling.
    \item \textbf{Scalability and Adaptive Computation}: Our initial results with depth-wise LoRA are promising but preliminary. Future work could explore dynamic iteration depth, inspired by latent space reasoning \cite{geiping2025scaling}, sparsity \cite{tan-etal-2023-sparse}, and mixture-of-experts \cite{csordas2024moeut}.
\end{enumerate}

\section*{Acknowledgements}
The research work was conducted in the JC STEM Lab of Multimedia and Machine Learning funded by The Hong Kong Jockey Club Charities Trust.

% Need to use this format for acknowledgement; see Wechat history
% \section*{References}
\newpage
\bibliographystyle{plain}
\bibliography{reference}

%%%%%%%%%%%%%%%%%%%%%%%%%%%%%%%%%%%%%%%%%%%%%%%%%%%%%%%%%%%%
\newpage
\appendix
\section{Theoretical Justification of Motivation}
\label{section:justification}
We provide here a theoretical foundation for the objective in~\eqref{eq:mle ultimate form}, showing how it arises naturally from a maximum likelihood estimation (MLE) framework under mild assumptions.

Let $\xvector \in \mathbb{R}^d$ be a random vector (considered as a token in the context of Transformer) in a high-dimensional representation space with a probability distribution $p(\xvector)$. Let $\{\zvector^h\}_{h=1}^H$ be a set of random vectors in low-dimensional latent spaces $\mathbb{R}^{p} ~(p<d)$ with distinct support, following a prior joint probability distribution $p(z^1, \dots, \zvector^H)$. We formulate information processing in the forward pass of neural networks as maximum likelihood estimation:
\begin{equation}
\label{eq:mle}
    \max_{\xvector} \quad \mathbb{E}_{(\zvector^1, \dots,\zvector^H)\sim p(\zvector^1,\dots,\zvector^H)}\left[\log p(\xvector,\zvector^1,\dots,\zvector^H; \theta,\phi)\right],
\end{equation}
where $\theta$ and $\phi$ are parameters of the high- and low-dimensional encodings, respectively.

To make this optimization more tractable, we make the following basic and practical assumptions:
\begin{assumption}
\label{assumption:1}
The random vectors $\zvector^1,\dots,\zvector^H$ are independent and follow the identical distribution $p(\zvector)$ in distinct latent spaces, \ie, $p(\zvector^1,\dots,\zvector^H)=\Pi_{h=1}^H p(\zvector^h)$ and $p(\zvector^1)=\dots=p(\zvector^H)=p(\zvector)$.
\end{assumption}

\begin{assumption}
\label{assumption:2}
The prior distribution $p(\zvector)$ is a uniform distribution with support on a hypersphere $\mathbb{S}^{p-1}$.
\end{assumption}

\begin{assumption}
\label{assumption:3}
The random vectors $(\zvector^1,\dots,\zvector^H)\sim p_{\phi}(\zvector^1,\dots,\zvector^H |\xvector)$ from the posterior distribution are conditionally independent, \ie, $p_{\phi}(\zvector^1,\dots,\zvector^H |\xvector) = \Pi_{h=1}^H p_\phi(\zvector^h|\xvector)$.
\end{assumption}

Assumption~\ref{assumption:2} of hyperspherical uniform distribution can be perceived to function as regularization on the latent representations to preserve maximum entropy and avoid representational collapse, which has been adopted to enhance auto-encoding \cite{xu-durrett-2018-spherical,s-vae18}. Under the above basic and practical assumptions, the MLE objective can be reformulated as:
\begin{align}
\label{eq:mle reformulation}
    \max_{\xvector} \quad & \mathbb{E}_{(\zvector^1, \dots,\zvector^H)\sim p(\zvector^1,\dots,\zvector^H)}\left[\log p(\xvector,\zvector^1,\dots,\zvector^H; \theta,\phi)\right] \nonumber \\
    & = \mathbb{E}_{(\zvector^1,\dots,\zvector^H) \sim p(\zvector)}\left[ \log p_\phi(\zvector^1,\dots,\zvector^H|\xvector)\right] + \mathbb{E}_{(\zvector^1,\dots,\zvector^H) \sim p(\zvector)} \left[\log p_\theta(\xvector) \right]  \nonumber\\
    & = \sum_{h=1}^H \mathbb{E}_{\zvector^h \sim p(\zvector)} \left[\log p_\phi(\zvector^h|\xvector) \right] + \log p_\theta(\xvector) \nonumber \\
    & = \sum_{h=1}^H \mathbb{E}_{\zvector^h \sim p(\zvector)} \left[\log \frac{p_\phi(\zvector^h|\xvector)}{p(\zvector^h)} \right] + \sum_{h=1}^H \mathbb{E}_{\zvector^h \sim p(\zvector)} \left[\log  p(\zvector^h) \right] + \log p_\theta(\xvector) \nonumber \\
    & = \sum_{h=1}^H \mathbb{E}_{\zvector^h \sim p(\zvector)} \left[\log \frac{p_\phi(\zvector^h|\xvector)}{p(\zvector)} \right] + \sum_{h=1}^H \mathbb{E}_{\zvector \sim p(\zvector)} \left[\log p(\zvector) \right] + \log p_\theta(\xvector) \nonumber \\
    & = -\sum_{h=1}^H \operatorname{D_{KL}}( p(\zvector) \parallel p_\phi(\zvector^h | \xvector) ) - H \times \mathcal{H}(p(\zvector)) +  \log p_\theta(\xvector), 
\end{align}
where $\operatorname{D_{KL}}(\cdot \parallel \cdot)$ denotes Kullback-Leibler (KL) divergence and $\mathcal{H}(\cdot)$ means differential entropy. As the second term on entropy in~\eqref{eq:mle reformulation} does not depend on variable $\xvector$, this objective ultimately reduces to~\eqref{eq:mle ultimate form} which we restate here for completeness:
\begin{equation*}
% \label{eq:mle ultimate form}
\min_{\xvector} \quad \sum_{h=1}^H \underbrace{\operatorname{D_{KL}}( p(\zvector) \parallel p_\phi(\zvector^h | \xvector) )}_{\text{uniformity}} \underbrace{- \log(p_\theta(\xvector))}_{\text{alignment}},
\end{equation*}

The first term encourages the posterior $p_\phi(\zvector^h|\xvector)$ defined on the vector $\zvector^h \in \mathbb{R}^p$ in latent space, which can be implemented by a transformation parameterized by $\phi$, to approximate a uniform distribution on a hypersphere. To see why the second term implies alignment, suppose the distribution $p_\theta(\xvector)$ is parameterized as a mixture of $M$ von Mises–Fisher (vMF) distributions \footnote{\href{https://en.wikipedia.org/wiki/Von_Mises–Fisher_distribution}{\text{https://en.wikipedia.org/wiki/Von\_Mises–Fisher\_distribution}}} with equal mixing coefficients:
\begin{equation}
\label{eq:vMF}
-\log(p_\theta(\xvector)) = -\log\left(\frac{1}{M} \sum_{m=1}^M f(\xvector; \muvector_m, \kappa_m) \right)= -\log\left(\frac{1}{M} \sum_{m=1}^M C_{d}(\kappa_m)\exp(\kappa_m \muvector_m^T\xvector)\right),
\end{equation}
where $\muvector_m \in \mathbb{S}^{d-1}$ denotes mean direction on $(d-1)$-dimensional unit sphere and $\kappa_m \geq 0$ is the concentration parameter while $C_{d}(\kappa_m)$ is the normalization constant; therefore, finding the vector $\xvector$ that minimizes this negative log-probability~\eqref{eq:vMF} equals finding maximal inner product $\muvector_m^T\xvector$, thus aiming for directional alignment. In practice, the mean direction $\muvector_m$ is learned by backpropagation and consequently contains certain statistical properties from the data. 

In summary, the objective~\eqref{eq:mle ultimate form} suggests that for a representation vector $\xvector \in \mathbb{R}^d$, the forward dynamics can be characterized by two complementary properties: 
\begin{itemize}[leftmargin=*]
    \item \textbf{Mode Seeking}: Achieving semantic alignment with directional vectors encapsulating specific information derived from data in the \textbf{high-dimensional space}.
    \item \textbf{Mass Covering}: Maximally preserving the entropy embedded via regularizing distributional uniformity in the \textbf{low-dimensional space}.
\end{itemize}
These principles underpin our design of token dynamics, and we propose to use energy functions to quantify these two properties as instantiations that can induce various Transformer-based models. 
% 1) seeking alignment with a set of directions in the \textbf{high-dimensional space} that encapsulate specific information derived from data per se, i.e. \textbf{mode seeking}; 2) maximally preserving the entropy embedded in the \textbf{low-dimensional space} via regularizing distribution uniformity, i.e., \textbf{mass covering}.  
% \todo{review}

\section{Preliminaries}
\label{section:preliminaries}
\subsection{Hopfield Networks}
Given a network with $N$ neurons $\xvector=[x_1, \dots, x_N]$ that take binary values, the temporal evolution dynamics of these neurons are determined by a scalar-value energy function:
\begin{equation*}
    E = - \frac{1}{2}\sum_{i,j}\omega_{ij}x_ix_j = -\frac{1}{2}\xvector^T\bm{W}\xvector,\quad x_{i}, x_{j} \in \{+1, -1\}
\end{equation*}
where $\omega_{ij}$ represents the strength of connectivity between node $x_i$ and $x_j$, and the connectivity is assumed to be symmetric, \ie, $\omega_{ij} = \omega_{ji}$. We can further rewrite $\bm{W} =\sum_{i=1}^P \bm{\xi}_i \bm{\xi}_i^T $ as a set of patterns to be stored. The update rule of each node to retrieve the most relevant pattern follows the Hebbian learning rule used in neuroscience: 
\begin{equation*}
    \xvector_{t+1} = \text{sign}(\bm{W}\xvector_t) = \text{sign}\left(\sum_{i=1}^P \bm{\xi}_i \bm{\xi}_i^T\xvector_t\right).
\end{equation*}

This update rule tends to minimize the energy function with retrieved patterns as its attractor. It is an embodiment of the idea of ``Neurons that fire together wire together.": If two neurons connect ($\omega_{ij}>0$), then they should have the same state ($+1$ for active and $-1$ for dead). The number of patterns the network can store and retrieve is $\mathcal{O}(N)$.

\subsection{Modern Continuous Hopfield Networks}
To overcome the limitation of linear storage capacity, modern Hopfield networks, also known as Dense Associative Memory \cite{krotov2016dense}, introduce nonlinearity in the energy and the update of neurons' states and make them suitable for continuous variables:
\begin{equation*}
    E =  -\frac{1}{2}\sum_{i=1}^Pf\left(\bm{\xi}_i^T\xvector\right),\ 
    \xvector_{t+1} = \operatorname{tanh}\left(\sum_{i=1}^P \bm{\xi}_i f'\left(\bm{\xi}_i^T\xvector_t\right)\right),
\end{equation*}
where $\operatorname{tanh(\cdot)}$ is to ensure the neurons' states are constrained to the interval $[-1,1]$ so that the energy is bounded from below. Depending on the form of $f$, the network could have power or exponential storage capacity. If we set $f(x) = x^2$, this reduces to the traditional networks with linear capacity. 

If we further make modifications to the non-linearity in the energy function with $\operatorname{logsumexp(\cdot)}$, which is inspired by contrastive normalization, we can define the Modern Continuous Hopfield (MCH) energy function with a quadratic regularization term on $\xvector$:
\begin{equation}
\label{eq:mch}
    E_{\text{MCH}} =  -\log\left(\sum_{i=1}^P \exp\left(\bm{\xi_i}^T\xvector\right)\right) + \frac{1}{2}\xvector^T\xvector.
\end{equation}

By leveraging the concave-convex procedure \cite{yuille2003concave}, the update could be written as  
\begin{equation*}
    \xvector_{t+1} =\Xi\operatorname{softmax}(\Xi^T\xvector_t),
\end{equation*}
where $\Xi = [\xi_1, \dots, \xi_P] \in \mathbb{R}^{N\times P}$. This formulation has proven to converge to stationary points of the energy function $E_{\text{MCH}}$, and is linked to the key-value memory similar to the attention mechanism \cite{ramsauer2021hopfield}. 
Notice that this update rule is essentially the cross-attention given a query vector $\xvector$ and can only describe the independent evolution of that vector. It fails to faithfully cover the parallel interactions between contextual tokens in the self-attention adopted in the GPT or BERT style Transformers. 

The construction of the modern continuous Hopfield energy and update rule can also be carried out from a biologically plausible view by extending the network with hidden neurons and establishing a group of coupled differential equations. We refer the readers to \cite{krotov2021large,krotov2023new} for more details.

\section{Derivation}
\label{appendix:B}

\subsection{Derivation of the Gradient of $E_{\text{ATTN}}$}
\label{section:derivation of E attn}
% \begin{equation}
% \begin{split}
\begin{align*}
\dot{\xvector}_k &= -\nabla_{\xvector_k} E_{\text{ATTN}} \\
&= -\sum_{h=1}^H\left(\frac{\sum_{j=1}^N \Wmatrix_h \Wmatrix_h^T \xvector_j \exp\left(\beta(\Wmatrix_h^T\xvector_k)^T(\Wmatrix_h^T\xvector_j)\right)}{\sum_{j=1}^N\exp\left(\beta (\Wmatrix_h^T\xvector_k)^T(\Wmatrix_h^T\xvector_j) \right)} + \sum_{i=1}^N \frac{\Wmatrix_h\Wmatrix_h^T\xvector_i\exp\left(\beta (\Wmatrix_h^T\xvector_i)^T(\Wmatrix_h^T\xvector_h) \right)}{\sum_{j=1}^N\exp\left(\beta (\Wmatrix_h^T\xvector_i)^T(\Wmatrix_h^T\xvector_j)\right)} \right) \\
& =-\sum_{h=1}^H \left(\Wmatrix_h\Wmatrix_h^T[\xvector_1, \dots,\xvector_N] \begin{bmatrix}
\exp\left(\beta (\Wmatrix_h^T\xvector_k)^T(\Wmatrix_h^T\xvector_1) \right) \\
\vdots \\
\exp\left(\beta (\Wmatrix_h^T\xvector_k)^T(\Wmatrix_h^T\xvector_N) \right) 
\end{bmatrix} / \sum_{j=1}^N\exp\left(\beta (\Wmatrix_h^T\xvector_k)^T(\Wmatrix_h^T\xvector_j) \right) + \right. \\
&\left.  \qquad \qquad \qquad \qquad \quad 
\sum_{i=1}^N \Wmatrix_h\Wmatrix_h^T\xvector_i\begin{bmatrix}
    \exp\left(\beta (\Wmatrix_h^T\xvector_1)^T(\Wmatrix_h^T\xvector_i) \right) / \sum_{j=1}^N\exp\left(\beta (\Wmatrix_h^T\xvector_i)^T(\Wmatrix_h^T\xvector_j)\right)\\
    \vdots \\
    \exp\left(\beta (\Wmatrix_h^T\xvector_N)^T(\Wmatrix_h^T\xvector_i) \right) / \sum_{j=1}^N\exp\left(\beta (\Wmatrix_h^T\xvector_i)^T(\Wmatrix_h^T\xvector_j)\right)
\end{bmatrix}_{k} \right)\\
&= -\sum_{h=1}^H \left(\Wmatrix_h\Wmatrix_h^T\Xmatrix\underbrace{\operatorname{softmax}}_{\text{column}}\left(\beta (\Wmatrix_h^T\Xmatrix)^T(\Wmatrix_h^T\xvector_k)\right) + \sum_{i=1}^N \Wmatrix_h \Wmatrix_h^T \xvector_i \underbrace{\operatorname{softmax}}_{\text{column}}\left(\beta (\Wmatrix_h^T\Xmatrix)^T(\Wmatrix_h^T\xvector_i)\right)_k\right) \\
&= -\sum_{h=1}^H \left(\Wmatrix_h\Wmatrix_h^T\Xmatrix\underbrace{\operatorname{softmax}}_{\text{column}}\left(\beta (\Wmatrix_h^T\Xmatrix)^T(\Wmatrix_h^T\xvector_k)\right) + \Wmatrix_h \Wmatrix_h^T \Xmatrix \underbrace{\operatorname{softmax}}_{\text{column}}\left(\beta (\Wmatrix_h^T\Xmatrix)^T(\Wmatrix_h^T\Xmatrix)\right)_{[k,:]}\right) \\
&= -\sum_{h=1}^H \left(\Wmatrix_h\Wmatrix_h^T\Xmatrix\underbrace{\operatorname{softmax}}_{\text{column}}\left(\beta (\Wmatrix_h^T\Xmatrix)^T(\Wmatrix_h^T\xvector_k)\right) + \Wmatrix_h \Wmatrix_h^T \Xmatrix \underbrace{\operatorname{softmax}}_{\text{row}}\left(\beta (\Wmatrix_h^T\Xmatrix)^T(\Wmatrix_h^T\Xmatrix)\right)_{[:,k]}\right) \\
&= -\sum_{h=1}^H \left(\Wmatrix_h\Wmatrix_h^T\Xmatrix\underbrace{\operatorname{softmax}}_{\text{column}}\left(\beta (\Wmatrix_h^T\Xmatrix)^T(\Wmatrix_h^T\xvector_k)\right) + \Wmatrix_h \Wmatrix_h^T \Xmatrix \underbrace{\operatorname{softmax}}_{\text{row}}\left(\beta (\Wmatrix_h^T\Xmatrix)^T(\Wmatrix_h^T\Xmatrix)\right)_{[:,k]}\right)
\end{align*}
% \end{split}
% \end{equation}

\begin{align*}
    \dot{\Xmatrix} &= [\dot{\xvector}_1, \dots, \dot{\xvector}_N] \\
                    &= -\nabla_{\Xmatrix} E_{\text{ATTN}}  \\
                    &= -\left( (\Wmatrix\Wmatrix^T\Xmatrix \underbrace{\operatorname{softmax}}_{\text{column-wise}}\left(\beta (\Wmatrix^T\Xmatrix)^T(\Wmatrix^T\Xmatrix)\right)  
+ \Wmatrix\Wmatrix^T\Xmatrix \underbrace{\operatorname{softmax}}_{\text{row-wise}}\left(\beta (\Wmatrix^T\Xmatrix)^T(\Wmatrix^T\Xmatrix)\right) \right) \\
\end{align*}

\subsection{Derivation of the Gradient of $E_{\text{FF}}$}
\label{section:derivation of E ff}

\begin{align*}
\dot{\xvector}_k &=  -\nabla_{\xvector_k}E_{\text{FF}} \\  
&=  \sum_{m=1}^M \operatorname{ReLU}(\dvector_m^T\xvector_k)\cdot\mathbb{I}(\dvector_m^T\xvector_k >0)\cdot \dvector_m \\
&= \sum_{m=1}^M \operatorname{ReLU}(\dvector_m^T\xvector_k)\dvector_m \\
&=[\dvector_1, \dots, \dvector_M]\begin{bmatrix}
    \operatorname{ReLU}(\dvector_1^T\xvector_k) \\
    \vdots \\
    \operatorname{ReLU}(\dvector_M^T\xvector_k)
\end{bmatrix}\\
&=\Dmatrix\operatorname{ReLU}(\Dmatrix^T\xvector_k)
\end{align*}

\begin{equation*}
\dot{\Xmatrix} = [\dot{\xvector}_1, \dots, \dot{\xvector}_N] = -\nabla_{\Xmatrix} E_{\text{FF}} = \Dmatrix\operatorname{ReLU}(\Dmatrix^T\Xmatrix)
\end{equation*}

\section{Detailed Experimental Setup and Model Configuration}
\label{section:detailed experimental setup}
\subsection{Network to Learn Adaptive Step Sizes}
\label{section:Network to Learn Adaptive Step Size}
\begin{figure}[htbp]
    \centering
    \includegraphics[width=\linewidth]{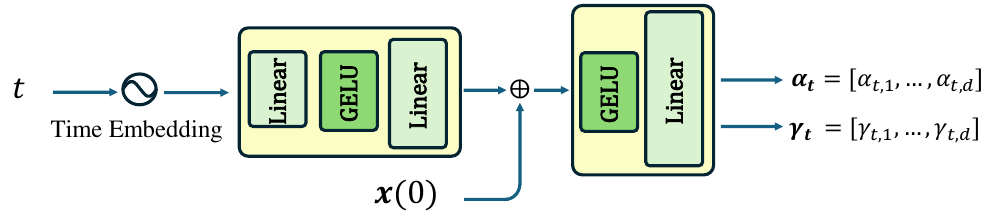}
        % \vskip -.2in
    \caption{Illustration of time embedding conditioned on the input to learn adaptive step size.}
    \label{fig:time embedding}
\end{figure}

% \todo{add some commets;and refer this figure on network}
We propose to learn adaptive step sizes $\bm{\alpha}_t,\bm{\gamma}_t \in \mathbb{R}^d$ given the current iteration $t \in [1,L]$, where $L$ is the iteration number of the layer with unique parameters, and the initial token $\xvector(0) \in \mathbb{R}^d$, using the network shown in Figure~\ref{fig:time embedding} and configurations in Table~\ref{tab:network to learn step sizes}.

\begin{table}[htbp]
\caption{Model configuration of network to learn adaptive step sizes.}
\label{tab:network to learn step sizes}
\centering
\begin{tabular}{ll}
        \toprule
        Layer  &  Configurations \\
        \midrule
        Time embedding  & 512 \\
        Linear & 512 $\times d$ \\
        GELU & -- \\
        Linear & $d \times d$ \\
        GELU & -- \\
        Linear & $d \times 2d$ \\
        \bottomrule
\end{tabular}
\end{table}

% \begin{figure*}[htbp]
%   \centering
%   \begin{minipage}{0.7\textwidth}
%     \centering
%     \includegraphics[width=\linewidth]{time_embedding.pdf} % Replace with your image
%     \caption{Illustration of time embedding conditioned on the input to learn adaptive step size.}
%     \label{fig:time embedding}
%   \end{minipage}
%   \hfill
%   \begin{minipage}{0.28\textwidth}
%   \caption{Model configuration of network to learn adaptive step sizes.}
%   \label{tab:network to learn step sizes}
%     \centering
%     \resizebox{\linewidth}{!}{
%     \begin{tabular}{ll}
%         \toprule
%         Layer  &  Configurations \\
%         \midrule
%         Time embedding  & 512 \\
%         Linear & 512 $\times d$ \\
%         GELU & -- \\
%         Linear & $d \times d$ \\
%         GELU & -- \\
%         Linear & $d \times 2d$ \\
%         \bottomrule
%     \end{tabular}
%     }
%     \label{tab:sample}
%   \end{minipage}
% \end{figure*}

\subsection{Solving Sudoku}
\label{section:sudoku setup}
Solving a Sudoku puzzle requires filling a 9x9 board, with some digits (1-9) known and unknown entries marked as 0. The unknown entries must be filled with digits perfectly such that the board satisfies a certain rule, which can be seen as a logical reasoning task \cite{wang2019satnet}. We tackle this puzzle by predicting the digits to fill in, conditioned on the given digits. 
%It can be viewed as a simplified masked modeling on synthetic data. 

Tables~\ref{tab:training recipe for solving Sudoku} and~\ref{tab:model configuration for solving Sudoku} show the training recipe and model configurations for solving Sudoku. We train the model with 24 iterations and can evaluate beyond these iterations.

\begin{table*}[htbp]
    \centering
    \begin{minipage}{0.49\textwidth}
    \centering
    \caption{Training recipe for solving Sudoku.}
    \label{tab:training recipe for solving Sudoku}
        % \resizebox{\linewidth}{!}{
        % \begin{tabular}{ll}
        \begin{tabularx}{\linewidth}{ll}
            \toprule
            Configuration  & Value \\
            \midrule
            Epochs  &  200 \\
            Batch size & 16 \\
            $\#$ GPU & 1 Nvidia 3090 \\
            $\#$ Training samples & 9k \\
            $\#$ Evaluating samples & 1k \\
            Optimizer & AdamW \\
            $\beta_1$,$\beta_2$ & 0.9, 0.95 \\
            Weight decay & 0.1  \\
            Learning rate (lr) & 1e-4\\
            Lr decay & Cosine \\
            Gradient clipping & 1.0 \\
            \bottomrule
        \end{tabularx}
        % \end{tabular}
        % }
    \end{minipage}%
    \hspace{\fill}
    \begin{minipage}{0.49\textwidth}
    \centering
    \caption{Model configuration for solving Sudoku.}
    \label{tab:model configuration for solving Sudoku} 
        % \resizebox{\linewidth}{!}{
        % \begin{tabular}{ll}
        \begin{tabularx}{\linewidth}{ll}
            \toprule
            Configuration  & Value \\
            \midrule
            Vocabulary size & 10 \\
            Layer  &  1 \\
            Iterations $L$ & 24 \\
            Hidden dimension $d$ & 768 \\
            Feedforward ratio $M$ & $4d$ \\
            Number of heads $H$ & 12 \\
            Positional encoding & Learnable \\
            Time embedding condition & $\Xmatrix_0$ \\
            Time embedding frequency & 512 \\
            \midrule
            Number of parameters & 5.20 M \\
            \bottomrule
        \end{tabularx}
        % \end{tabular}
        % }
    \end{minipage}
\end{table*}

\subsection{Image Classification}
\label{section:image classification setup}
Tables~\ref{tab:training recipe for image classfication} and~\ref{tab:model configuration for image classfication} present the training recipe and model configurations on image classification, where the number of our model is computed on CIFAR-10. In practice, we use absolute sinusoidal positional encoding and adopt conditioning on $\Xmatrix_t$ for performance reasons. Table~\ref{tab:scaling modeling configuration} lists the configuration of different sizes and it applies to other tasks as well.

\begin{table*}[htbp]
    \centering
    \begin{minipage}{0.49\textwidth}
        \centering
        \captionof{table}{Training recipe for image classification on CIFAR-10/100.}
        \label{tab:training recipe for image classfication}
        % \begin{tabular}{ll}
        \begin{tabularx}{\linewidth}{ll}
            \toprule
            Configuration  & Value \\
            \midrule
            Epochs  &  200 \\
            Batch size & 128 \\
            $\#$ GPU & 1 Nvidia 3090 \\
            $\#$ Training samples & 50k \\
            $\#$ Evaluating samples & 10k \\
            Optimizer & Adam \\
            $\beta_1$,$\beta_2$ & 0.9, 0.999 \\
            Weight decay & 5e-5  \\
            Max learning rate (lr) & 1e-3\\
            Min learning rate (lr) & 1e-5\\
            Lr decay & Cosine \\
            Warmup epochs & 5 \\
            Input size & 32 \\
            \bottomrule
        % \end{tabular}
        \end{tabularx}
    \end{minipage}%
    % \hspace{0.08\textwidth} % Space between the tables
    \hspace{\fill}
    \begin{minipage}{0.49\textwidth}
        \centering
        \captionof{table}{Model configuration for image classification on CIFAR-10/100.}
        \label{tab:model configuration for image classfication}
        % \begin{tabular}{ll}
        \begin{tabularx}{\linewidth}{ll}
            \toprule
            Configuration ($\texttt{Small}$) & Value \\
            \midrule
            Patch size & 8 \\
            Layer  &  1 \\
            Iterations $L$  & 12 \\
            Hidden dimension $d$ & 384 \\
            Feedforward ratio $M$ & $d$ \\
            Number of heads $H$ & 6 \\
            Positional encoding & Sinusoidal\\
            Time embedding condition & $\Xmatrix_t$ \\
            Time embedding frequency & 512 \\
            \midrule
            Number of parameters & 0.96 M \\
            \bottomrule
        % \end{tabular}
        \end{tabularx}
    \end{minipage}
\end{table*}

\begin{table*}[htbp]
    \centering
    \begin{minipage}{0.49\textwidth}
        \centering
        \captionof{table}{Training recipe for image classification on ImageNet-100.}
        \label{tab:training recipe for image classification im100}
        % \begin{tabular}{ll}
        \begin{tabularx}{\linewidth}{ll}
            \toprule
            Configuration  & Value \\
            \midrule
            Epochs  &  200 \\
            Batch size & 256 \\
            $\#$ GPU & 1 Nvidia A100 \\
            $\#$ Training samples & 126,689 \\
            $\#$ Evaluating samples & 5k \\
            Optimizer & Adam \\
            $\beta_1$,$\beta_2$ & 0.9, 0.999 \\
            Weight decay & 5e-5  \\
            Max learning rate (lr) & 1e-3\\
            Min learning rate (lr) & 1e-5\\
            Lr decay & Cosine \\
            Warmup epochs & 5 \\
            Input size & 224 \\
            \bottomrule
        % \end{tabular}
        \end{tabularx}
    \end{minipage}%
    % \hspace{0.08\textwidth} % Space between the tables
    \hspace{\fill}
    \begin{minipage}{0.49\textwidth}
        \centering
        \captionof{table}{Model configuration for image classification on ImageNet-100.}
        \label{tab:model configuration for image classification im100}
        % \begin{tabular}{ll}
        \begin{tabularx}{\linewidth}{ll}
            \toprule
            Configuration ($\texttt{Small}$) & Value \\
            \midrule
            Patch size & 16 \\
            Layer  &  1 \\
            Iterations $L$ & 12 \\
            Hidden dimension $d$ & 384 \\
            Feedforward ratio $M$ & $d$ \\
            Number of heads $H$ & 6 \\
            Positional encoding & Sinusoidal\\
            Time embedding condition & $\Xmatrix_t$ \\
            Time embedding frequency & 512 \\
            \midrule
            Number of parameters & 1.21 M \\
            \bottomrule
        % \end{tabular}
        \end{tabularx}
    \end{minipage}
\end{table*}

\begin{table}[htbp]
\caption{Model configuration of different sizes}
\label{tab:scaling modeling configuration}
\centering
\begin{tabular}{llll}
\toprule
Configurations  &  $\texttt{Small}$ & $\texttt{Small}$ Scale-up & $\texttt{Base}$\\
\midrule
Hidden dimension $d$  & 384 & 512 & 768 \\
Number of heads $H$  & 6 & 8 & 12 \\
\bottomrule
\end{tabular}

\end{table}

\subsection{Masked Image Modeling}
\label{section:masked image modeling setup}

We follow \cite{chang2022maskgit} using VQ-VAE to tokenize the images to $16\times16$ latent code with the codebook size of 1024 after resizing the input to $256\times256$. The masking ratio is randomly chosen between $[0,0.4]$, and the masked region is replaced by a learnable token. Training loss is computed only for the masked tokens. We also follow the iterative decoding process in \cite{chang2022maskgit} with $\text{temperature}=1$ and decoding step $T=24$. We also remove the MLP following the time embedding and set the embedding 
 frequency equal to the hidden dimension to save parameters, and we find out that this implementation works better. Tables~\ref{tab:training recipe for masked image modeling} and~\ref{tab:model configuration for masked image modeling} show the detailed training recipe and configurations.

 We evaluate the quality of the reconstructed images of masking out $40\%$ of the images with $\texttt{Base}$ configurations. We report Peak Signal-to-Noise Ratio (PSNR), Structural Similarity Index Measure (SSIM) \cite{wang2004image}, Multi-Scale SSIM \cite{wang2003multiscale}, Learned Perceptual Image Patch Similarity (LPIPS) \cite{zhang2018unreasonable} and Fr$\acute{\text{e}}$chet Inception Distance (FID) \cite{heusel2017gans} on the validation set (5k). 
\begin{table*}[htbp]
    \centering
    \begin{minipage}{0.49\textwidth}
    \caption{Training recipe for masked image modeling.}
    \label{tab:training recipe for masked image modeling}
        \centering
        % \begin{tabular}{ll}
        \begin{tabularx}{\linewidth}{ll}
            \toprule
            Configuration  & Value \\
            \midrule
            Epochs  &  300 \\
            Batch size & 256 (64 $\times$4)\\
            $\#$ GPU & 4 Nvidia 80 GB A100 \\
            $\#$ Training samples & 126,689 \\
            $\#$ Evaluating samples & 5,000 \\
            Optimizer & AdamW \\
            $\beta_1$, $\beta_2$ & 0.9, 0.95 \\
            Weight decay & 0.1  \\
            Learning rate (lr) & 1e-4\\
            Lr decay & None \\
            Gradient clipping & 3.0 \\
            Input size & 256 \\
            \bottomrule
        % \end{tabular}
        \end{tabularx}
    \end{minipage}%
    % \hspace{0.08\textwidth} % Space between the tables
    \hspace{\fill}
    \begin{minipage}{0.45\textwidth}
        \caption{Model configuration for masked image modeling.}
    \label{tab:model configuration for masked image modeling}
        \centering
        % \begin{tabular}{ll}
        \begin{tabularx}{\linewidth}{ll}
            \toprule
            Configuration  & Value \\
            \midrule
            Vocabulary size & 1025 \\
            Layer  &  1 \\
            Iterations & 12 \\
            Hidden dimension $d$ & 768 \\
            Feedforward ratio $M$ & $d$ \\
            Number of heads $H$ & 12 \\
            Positional encoding & Sinusoidal \\
            Time embedding condition & $\Xmatrix_0$ \\
            Time embedding frequency & 768 \\
            \midrule
            Number of parameters & 3.94 M \\
            \bottomrule
        % \end{tabular}
        \end{tabularx}
    \end{minipage}
\end{table*}

\subsection{Definition of Effective Rank and Average Angle}
\label{section:definition}
We provide the formal definition of effective rank~\eqref{eq:13} and average angle~\eqref{eq:14} below. The effective rank is a continuous proxy of the full rank and, similar to the average angle, reflects the extent to which a set of vectors distributes uniformly. 
\begin{definition}[Effective Rank]
\label{definition:effective rank}
For a matrix $\Xmatrix \in \mathbb{R}^{d\times N}$, let $\Sigma=[\sigma_1,\dots,\sigma_r]$ be its singular values where $r$ is its full rank and denote $p_i = \sigma_i / \sum_{j=1}^r \sigma_j$ the discrete probability. The effective rank \cite{roy2007effective,guo2023contranorm} is defined as the exponential of the entropy 
\begin{equation}
\label{eq:13}
\exp(-\sum_{i=1}^r p_i \log p_i).
\end{equation}
\end{definition}

\begin{definition}[Average Angle]
\label{definition:average angle}
Given a set of vectors $\Xmatrix = [\xvector_1, \dots, \xvector_N] \in \mathbb{R}^{d\times N}$, the average angle of these vectors is 
\begin{equation}
\label{eq:14}
\operatorname{arccos} \frac{2}{N(N-1)}\sum_{i=1}^N\sum_{j=i+1}^{N}\frac{\xvector_i^T\xvector_j}{\|\xvector_i\|_2\|\xvector_j\|_2}.
\end{equation}
\end{definition}

\section{Additional Results of Masked Image Modeling}
\label{section:additional results of masked image modeling}

\begin{figure}[htbp]
    \centering
    \includegraphics[width=\linewidth]{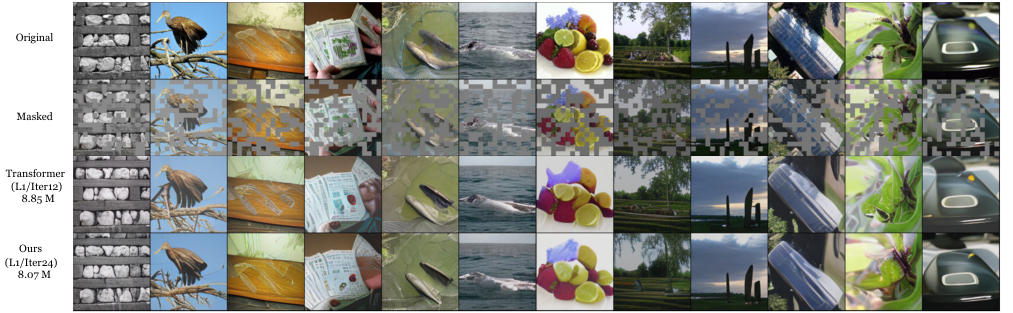}
    \includegraphics[width=\linewidth]{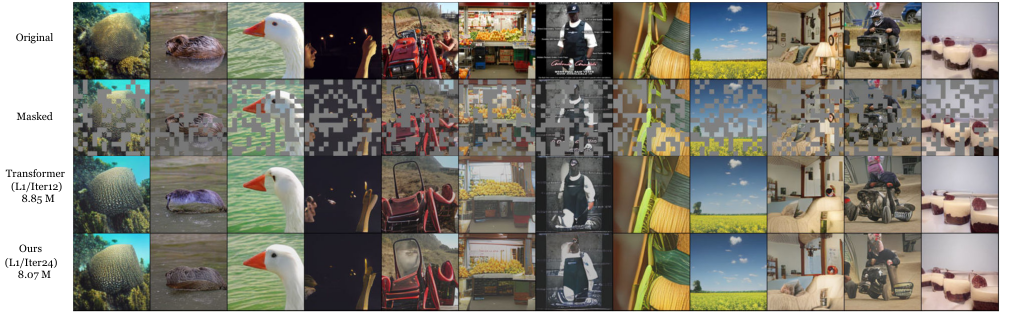}
    \includegraphics[width=\linewidth]{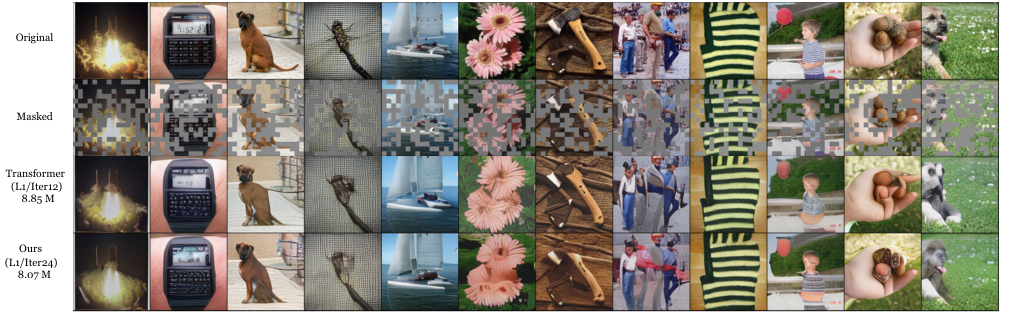}
    \caption{Visual comparison on masked image modeling on ImageNet 256$\times$256. Our model, when scaling to Transformer scale with additional compute, can achieve similar reconstruction quality when masking ratio $=40\%$.}
    \label{fig:visualization more results}
\end{figure}

Tables~\ref{tab:masked image modeling 0.1},~\ref{tab:masked image modeling 0.2}, and~\ref{tab:masked image modeling 0.3} summarize the results of masked image modeling with different masking ratios. When scaled to larger iterations and a wider feedforward module, our model achieves comparable results to Transformer but still slightly lags behind. This suggests the scalability of our model to the large configuration may be a bottleneck for its development and deployment. More visual comparison is provided in Figure~\ref{fig:visualization more results}.

\begin{table}[htbp]
\caption{Comparison of masked image modeling performance of masking ratio $10\%$. }
\label{tab:masked image modeling 0.1}
\centering
\resizebox{\linewidth}{!}{
\begin{tabular}{llllcll}
\toprule
Models  &  Layer / Iteration / FF Ratio $M$ (\# Params)& PSNR ($\uparrow$) & SSIM ($\uparrow$) & Multi-Scale SSIM ($\uparrow$) & LPIPS ($\downarrow$)& FID ($\downarrow$)\\
\midrule
Transformer  & L1 / Iter 12 / 4$d$ ~(8.85 M) & \textbf{17.693} & \textbf{0.466} & \textbf{0.709} & 0.236 & \textbf{22.428}
\\
Ours  & L1 / Iter 12 / $d$ ~(3.94 M) & 17.553 & 0.462 & 0.701 & 0.243 & 24.665\\
Ours  & L1 / Iter 24 / 8$d$ ~(8.07 M) & 17.673 & 0.465 & 0.708 & \textbf{0.236} & 22.517\\
\bottomrule
\end{tabular}
}
\end{table}

\begin{table}[htbp]
\caption{Comparison of masked image modeling performance of masking ratio $20\%$. }
\label{tab:masked image modeling 0.2}
\centering
\resizebox{\linewidth}{!}{
\begin{tabular}{llllcll}
\toprule
Models  &  Layer / Iteration / FF Ratio $M$ (\# Params)& PSNR ($\uparrow$) & SSIM ($\uparrow$) & Multi-Scale SSIM ($\uparrow$) & LPIPS ($\downarrow$)& FID ($\downarrow$)\\
\midrule
Transformer  & L1 / Iter 12 / 4$d$ ~(8.85 M) & \textbf{17.185} & \textbf{0.451} & \textbf{0.678} & \textbf{0.261} & \textbf{27.320}
\\
Ours  & L1 / Iter 12 / $d$ ~(3.94 M) & 16.988 & 0.444 & 0.662 & 0.275 & 33.637\\
Ours  & L1 / Iter 24 / 8$d$ ~(8.07 M) & 17.170 & 0.450 & 0.676 & 0.262 & 28.120\\
\bottomrule
\end{tabular}
}
\end{table}

\begin{table}[htbp]
\caption{Comparison of masked image modeling performance of masking ratio $30\%$. }
\label{tab:masked image modeling 0.3}
\centering
\resizebox{\linewidth}{!}{
\begin{tabular}{llllcll}
\toprule
Models  &  Layer / Iteration / FF Ratio $M$ (\# Params)& PSNR ($\uparrow$) & SSIM ($\uparrow$) & Multi-Scale SSIM ($\uparrow$) & LPIPS ($\downarrow$)& FID ($\downarrow$)\\
\midrule
Transformer  & L1 / Iter 12 / 4$d$ ~(8.85 M) & \textbf{16.616} & \textbf{0.435} & \textbf{0.642} & \textbf{0.291 }& \textbf{35.095}
\\
Ours  & L1 / Iter 12 / $d$ ~(3.94 M) & 16.365 & 0.427 & 0.621 & 0.314 & 45.642\\
Ours  & L1 / Iter 24 / 8$d$ ~(8.07 M) & 16.590 & 0.434 & 0.638 & 0.294 & 35.128\\
\bottomrule
\end{tabular}
}
\end{table}

\newpage
\section{Rank and Average Angle of Each Head}
\label{section:rank angle sudoku cifar10 complete}
\subsection{Sudoku Dataset}
\label{section:rank angle sudoku complete}

\begin{figure}[htbp]
    \centering
    \includegraphics[width=\linewidth]{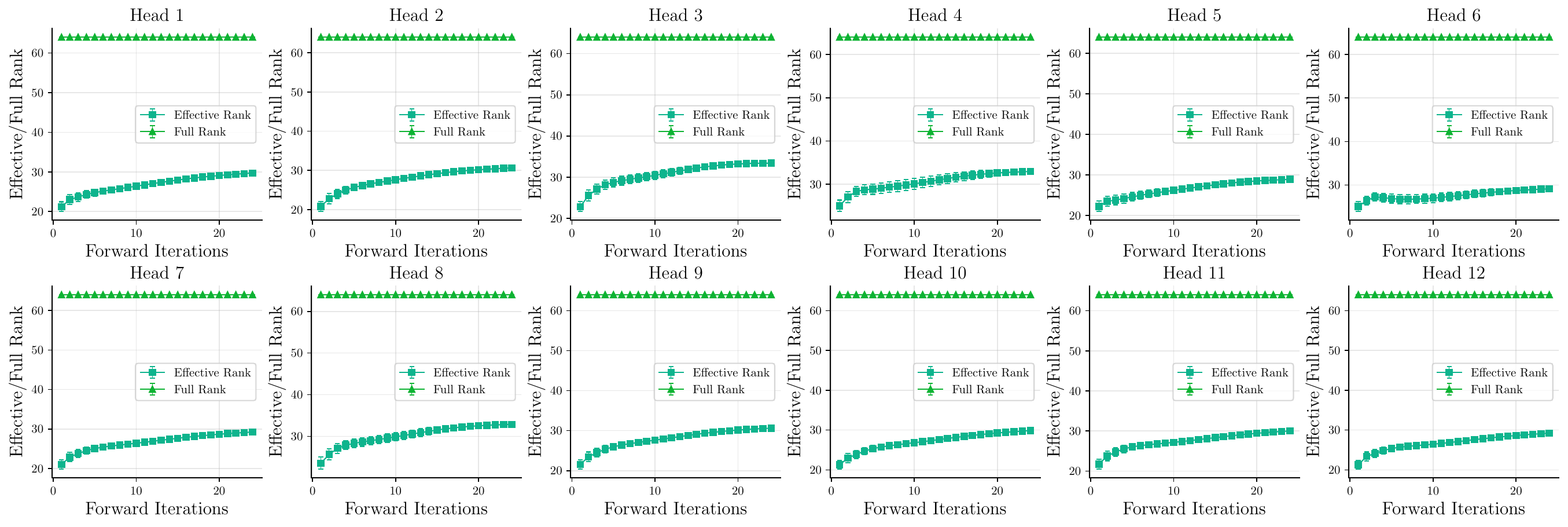}
    \caption{The effective rank of tokens projected to each subspace. Results are from the test set of Sudoku dataset \cite{palm2018recurrent}.}
    \label{fig:rank sudoku complete}
\end{figure}

\begin{figure}[htbp]
    \centering
    \includegraphics[width=\linewidth]{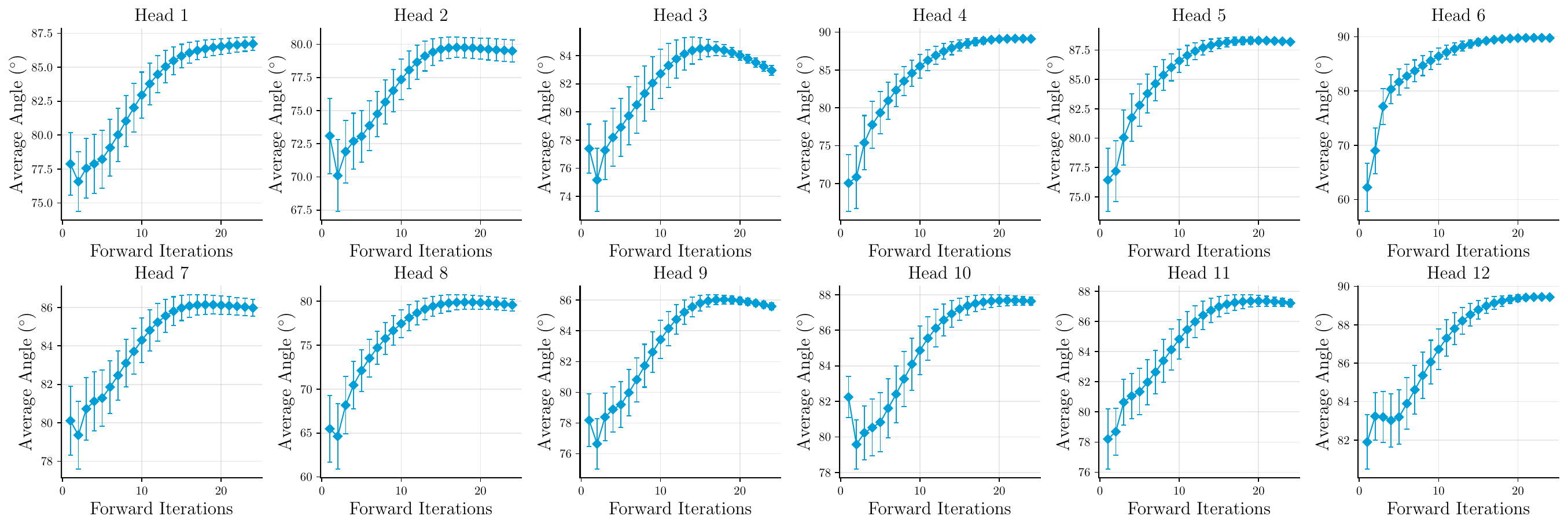}
    \caption{The average angle of tokens projected to each subspace. Results are from the test set of Sudoku dataset \cite{palm2018recurrent}.}
    \label{fig:angle sudoku complete}
\end{figure}

Figures~\ref{fig:rank sudoku complete} and~\ref{fig:angle sudoku complete} capture the evolution of the effective rank and average angle of all heads. Most of them follow the separation dynamics on the hypersphere where tokens tend to be near-orthogonal, corroborating our design goal of attention energy. 

%%%%%%%%%%%%%%%%%%%%%%%%%%%%%%%%%%%%%%%%%%%%%%%%%%%%%%%%%%%%%%%%%%%%%%%%%%%%%%%
%%%%%%%%%%%%%%%%%%%%%%%%%%%%%%%%%%%%%%%%%%%%%%%%%%%%%%%%%%%%%%%%%%%%%%%%%%%%%%%
\begin{figure}[htbp]
    \centering
    \includegraphics[width=\linewidth]{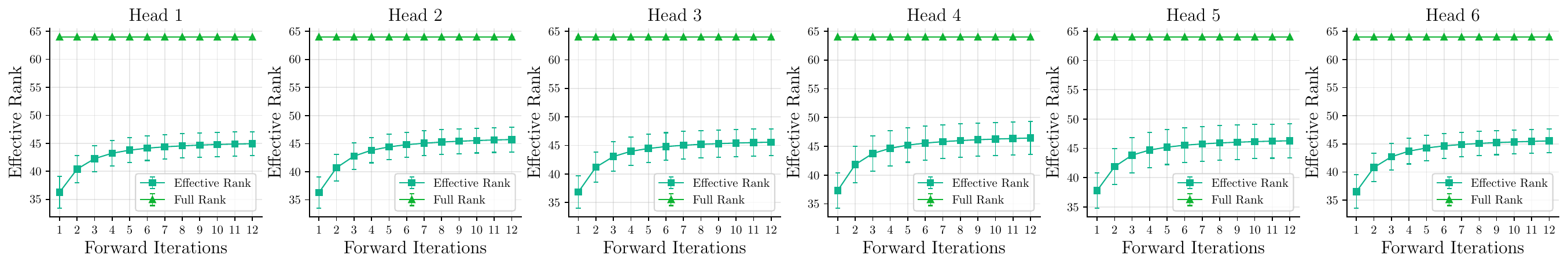}
    \caption{The effective rank of tokens projected to each subspace. Results are from CIFAR-10 validation set.}
    \label{fig:rank cifar10 complete}
\end{figure}

\begin{figure}[htbp]
    \centering
    \includegraphics[width=\linewidth]{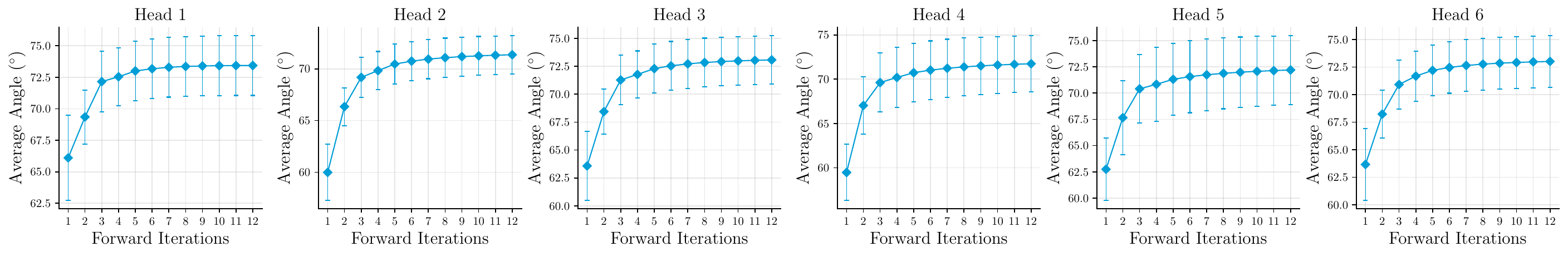}
    \caption{The average angle of tokens projected to each subspace. Results are from CIFAR-10 validation set.}
    \label{fig:angle cifar10 complete}
\end{figure}

\begin{figure}[htbp]
    \centering
    \includegraphics[width=\linewidth]{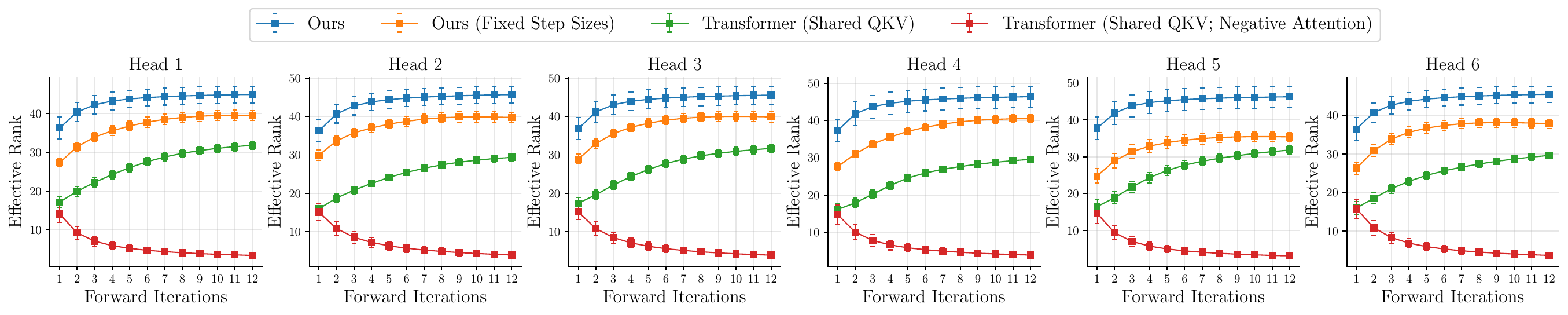}
    \caption{Comparison of the effective rank at each subspace with fixed step sizes, Transformer with shared query/key/value matrix, and reverting the update sign before its attention. Results are from CIFAR-10 validation set.}
    \label{fig:rank shared QKV cifar10 complete}
\end{figure}

\begin{figure}[htbp]
    \centering
    \includegraphics[width=\linewidth]{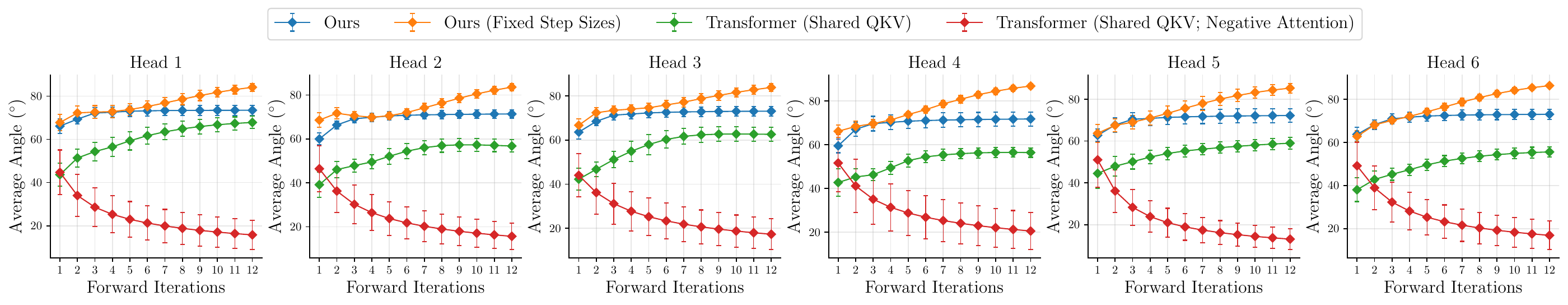}
    \caption{Comparison of the average angle at each subspace with fixed step sizes, Transformer with shared query/key/value matrix, and reverting the update sign before its attention. Results are from CIFAR-10 validation set.}
    \label{fig:angle shared QKV cifar10 complete}
\end{figure}

\subsection{CIFAR-10 Dataset}
\label{section:rank angle cifar10 complete}
The full results on CIFAR-10 also possess similar trends to those on the Sudoku dataset, as shown in Figures~\ref{fig:rank cifar10 complete} and~\ref{fig:angle cifar10 complete}.

\subsection{Comparison with Transformer with Shared Query, Key and Value Matrix}

A notable connection between Hyper-SET and vanilla Transformer lies in the shared query (Q), key (K), and value (V) projection matrix, which has been studied recently \cite{kowsher-etal-2025-self}. To verify whether Hyper-SET captures essential Transformer behaviors, we adapt vanilla Transformer to have shared QKV projections and measure its effective rank and average angle among projected tokens. Furthermore, we also include comparisons with Hyper-SET that adopts fixed step sizes set as 0.1 to evaluate the necessity of learned ones.

As shown in Figures~\ref{fig:rank shared QKV cifar10 complete} and~\ref{fig:angle shared QKV cifar10 complete}, both Hyper-SET and its fixed-step variant exhibit increasing token separation across subspaces, confirming the emergence of distributional uniformity and the benefit of learned step sizes. This dynamics is also mirrored in shared QKV Transformer, which cross-validates our insights on distributional uniformity in subspaces, suggesting the promise of this parameter-sharing design. In contrast, modifying the shared Transformer to reverse the update direction of attention—similar to the design in~\eqref{eq:8}—leads to a decline in both rank and angle, highlighting a breakdown in uniformity. This contrast emphasizes that Hyper-SET is not a heuristic tweak of vanilla Transformer but a principled architecture.

\section{Additional Results on Alternative Designs and Scalability}
\label{section:alternative designs and scalability}
\subsection{Alternative Designs on Energy Functions}
\label{section:alternative designs}
Our proposed energy functions $E_\text{ATTN}$ and $E_\text{FF}$ provide an avenue to quantify the objective on uniformity and alignment in an amenable way for optimization. To manifest the significance of our conceptualization in designing a variety of Transformer-based models, we extend the energy functions to more general forms and provide alternative instantiations of them that can induce novel structures. Specifically, we generalize the energy functions in~\eqref{eq:2} and~\eqref{eq:4} to the following forms:
\begin{align}
E_\text{ATTN} &= \sum_{h=1}^H E_\text{ATTN}^h = \sum_{h=1}^H\sum_{i=1}^N f\left( \sum_{j=1}^N K(\Wmatrix_h^T\xvector_i, \Wmatrix_h^T\xvector_j)\right), \label{eq:generalized attn} \\
E_\text{FF} &= - \sum_{i=1}^N g\left(\sum_{m=1}^M h(\dvector_m^T \xvector_i)\right). \label{eq:generalized ff}
\end{align}
where $K:\mathbb{R}^p \times \mathbb{R}^p \rightarrow \mathbb{R} $ is a kernel function and $f,g,h: \mathbb{R} \rightarrow\mathbb{R}$ are non-decreasing scalar functions. For clarity, we omit the hyperspherical constraint but its correspondence to $\operatorname{RMSNorm}$ still holds. 

By choosing different variations on these functions, we arrive at alternative energy functions and their consequent attention and feedforward operators, of which we summarize the specifications in the following tables. Remarkably, in Table~\ref{tab:generalized attention}, we can derive an attention with linear complexity $\mathcal{O}(N)$ by specifying the kernel function with the inner product of an element-wise transformation $\Phi: \mathbb{R} \rightarrow \mathbb{R}$, bridging our energy view with recent advances in linear attention design. In practice, we choose it as the sigmoid function $\sigma(x) = 1 /(1+\exp(-x))$, but other designs can also be possible. In Table~\ref{tab:generalized feedforward}, if we specify the outer function $g$ in the feedforward energy as a quadratic function, there is a novel summation and a Hadamard product operation emerging with the transformation $\Phi$, similar to the gating mechanism. We also specify $\Phi$ with the sigmoid function.

\begin{table}[tbp]
\caption{Alternative designs on attention energy $E_\text{ATTN}$ and the induced operators.}
\label{tab:generalized attention}
\centering
\resizebox{\linewidth}{!}{
\begin{tabular}{lllll}
\toprule
% \multirow{2}{*}{Models}  & \multirow{2}{*}{Config ($\#$ Params)} & \multicolumn{3}{c}{Dataset}\\
% \cmidrule(lr){3-5}
%  &  & CIFAR-10 & CIFAR-100 & IM-100\\
Operator & $f(x)$ & $K(\bm{x},\bm{y})$ & $E_\text{ATTN}$ & $-\nabla_{\Xmatrix} E_\text{ATTN}$ \\
\midrule
Bi-Softmax (Ours) & $\beta^{-1}\log(x)$ & $\exp(\beta \bm{x}^T\bm{y})$ & \eqref{eq:2} & \eqref{eq:6} \\
Sigmoid-Softmax  & $\frac{\beta^{-1}}{2}x$ & $\sigma(\beta \bm{x}^T\bm{y})$ & $\frac{1}{2}\sum\limits_{h=1}^H\sum\limits_{i=1}^N\sum\limits_{j=1}^N\sigma\left(\beta (\Wmatrix_h^T\xvector_i)^T\Wmatrix_h^T\xvector_j\right) \beta^{-1}$ & $\sum\limits_{h=1}^H \Wmatrix_h \Wmatrix_h^T \Xmatrix \sigma(1-\sigma)\left( \beta (\Wmatrix_h^T \Xmatrix)^T \Wmatrix_h^T \Xmatrix\right)$ \\
Linear Attention & $\frac{\beta^{-1}}{2}(x)$ & $\frac{1}{2}\left(\beta\Phi(\bm{x})^T\Phi(\bm{y})\right)^2$ & $\frac{1}{4}\sum\limits_{h=1}^H\sum\limits_{i=1}^N\sum\limits_{j=1}^N\left( \beta \Phi(\Wmatrix_h^T\xvector_i)^T \Phi(\Wmatrix_h^T\xvector_j)\right)^2 \beta^{-1}$ & $\sum\limits_{h=1}^H \Wmatrix_h \Phi^{\prime}(\Wmatrix_h^T\Xmatrix) \odot \left(\beta \Phi(\Wmatrix_h^T\Xmatrix)\Phi(\Wmatrix_h^T\Xmatrix)^T\Phi(\Wmatrix_h^T\Xmatrix)\right)$ \\
\bottomrule
\end{tabular}
}
\end{table}

\begin{table}[tbp]
\caption{Alternative designs on feedforward energy $E_\text{FF}$ and the induced operators.}
\label{tab:generalized feedforward}
\centering
\resizebox{\linewidth}{!}{
\begin{tabular}{lllll}
\toprule
% \multirow{2}{*}{Models}  & \multirow{2}{*}{Config ($\#$ Params)} & \multicolumn{3}{c}{Dataset}\\
% \cmidrule(lr){3-5}
%  &  & CIFAR-10 & CIFAR-100 & IM-100\\
Operator & $g(x)$ & $h(x)$ & $E_\text{FF}$ & $-\nabla_{\Xmatrix} E_\text{FF}$ \\
\midrule
ReLU FF (Ours) & $x$ & $\frac{1}{2}\operatorname{ReLU}^2(x)$ & \eqref{eq:4} & \eqref{eq:9} \\
Softmax FF & $\log(x)$ & $\exp(x)$ & $-\sum\limits_{i=1}^N\log\left(\sum\limits_{m=1}^M\exp(\dvector_m^T\xvector_i) \right)$ & $\Dmatrix \underbrace{\operatorname{softmax}}_{\text{column-wise}}(\Dmatrix^T \Xmatrix)$ \\
Gated FF & $\frac{1}{2}x^2$ & $\Phi(x)$ & $-\frac{1}{2} \sum\limits_{i=1}^N\left(\sum\limits_{m=1}^M \Phi\left(\dvector_m^T \xvector_i \right) \right)^2$ &  $\Dmatrix \underbrace{\Phi(\Dmatrix^T\Xmatrix)}_{\text{column sum}} \odot ~\Phi^{\prime}(\Dmatrix^T\Xmatrix)$\\
\bottomrule
\end{tabular}
}
\end{table}

In summary, these design choices demonstrate that Hyper-SET is more than just a single model, but a blueprint for constructing principled Transformer variants. Each component—self-attention, feedforward, and normalization—can be systematically interpreted and designed within our energy minimization framework, providing a pathway for principled, modular innovation in sequence model architectures. 
\subsection{Exploratory Results on Scalability}
\label{section:scalability}
% \todo{This design maintains the core recurrence-based parameter sharing of Hyper-SET while allowing for depth-aware expressiveness, enabling better adaptation to increasingly complex tasks as the model scales. ; both AB are initialized as Gaussain with 0.02 std; scaling factor 4}

\begin{table}[htbp]
\caption{Top-1 accuracy on image classification when scaling up the model size. Our model surpasses other baselines while being parameter-efficient compared to vanilla Transformer.}
\label{tab:scaling up}
\centering
% \resizebox{\linewidth}{!}{
\begin{tabular}{lllll}
\toprule
% \multirow{2}{*}{Models}  & \multirow{2}{*}{Config ($\#$ Params)} & \multicolumn{3}{c}{Dataset}\\
% \cmidrule(lr){3-5}
%  &  & CIFAR-10 & CIFAR-100 & IM-100\\
Dataset & Models & Width $d$ & $\#$ Params & Accuracy (\%) \\
\midrule
\multirow{4}{*}{CIFAR-10} & Transformer & 1152 & 16.0 M & 89.42 \\
                          & CRATE \cite{yu2023white} & 1152 & 4.1 M & 88.77 \\
                          & Energy Transformer \cite{hoover2024energy} & 1152 & 8.0 M & 76.21 \\
                          & Ours & 1536 & 14.3 M & \textbf{90.62} \\
\midrule
\multirow{4}{*}{CIFAR-100} & Transformer & 1152 & 16.1 M & 62.83 \\
                          & CRATE \cite{yu2023white} & 1152 & 4.2 M & 63.39 \\
                          & Energy Transformer \cite{hoover2024energy} & 1152 & 8.1 M & 55.47 \\
                          & Ours & 1536 & 14.4 M & \textbf{66.30} \\
\bottomrule
\end{tabular}
% }
\end{table}

To equip our model with flexible computation while maintaining its core recurrence-based parameter sharing, we add an independent low-rank adaptation \cite{hu2022lora} matrix $W=AB$ to every iteration of the Transformer layer sharing the same base parameters. Inspired by but unlike depth-wise adaptation of pre-trained models in \cite{bae2025relaxed}, we train the base parameters and the adaptation matrix together. Both matrix $A \in \mathbb{R}^{d\times r},B \in \mathbb{R}^{r\times d}$ with rank $r$ are initialized with Gaussain of 0.02 standard deviation. The scaling factor before the adaptation matrix is set as 4.

To further demonstrate the capability of our model when scaling up its size, we perform a preliminary evaluation on image classification. Following the experimental setup in the main paper to configure the models with one layer and repeat with 12 iterations, we scale up the width 
 of Transformer to 1152 and ours to 1536, resulting in similar parameter size. The result is presented in Table~\ref{tab:scaling up}. These explorations showcase the potential of Hyper-SET in improving model capacity without significantly increasing the total parameter count.

% \section{Discussion and Limitations}
\section{Discussion and Broader Impact}
\label{section:discussion and limitations}
% \todo{add discussion}
\subsection{Connections and Differences with Energy Transformer}
Both our Hyper-SET and ET~\cite{hoover2024energy} are grounded in the idea of interpreting Transformer components as gradient flows that minimize an energy function. However, the two approaches diverge significantly in motivations, theoretical formulation, and architectural design.

% \paragraph{Motivational Differences}
% Our work explicitly targets the token synchronization problem by combining uniformity and alignment objectives (Section~\ref{section:energy functions}). The Hopfield-style attention energy in Hyper-SET serves as a specific instantiation of a more general principle: minimizing hyperspherical interaction to prevent token collapse.

% In contrast, ET primarily uses energy dynamics as an alternative lens for interpreting softmax attention, but does not directly connect the energy formulation to any particular representational challenge, such as token collapse or generalization. Moreover, ET adopts Hopfield energy more as a starting point than as a motivation-driven design.

\begin{itemize}[leftmargin=*]
    \item \textbf{Motivation}
        \begin{itemize}[leftmargin=*]
            \item \textbf{Hyper-SET}: centers on a dual objective of semantic alignment (mode seeking) and uniformity (mass covering) under hyperspherical constraints grounded in maximal likelihood estimation. The proposed Hopfield-style energy aims to quantify it into an optimizable objective and serves as a specific instantiation of this more general principle, which fundamentally differs from ET.
            \item \textbf{ET}: maintains the mechanistic interpretation of associative memory and does not directly connect the energy formulation to any particular representational challenge. Moreover, ET adopts Hopfield energy more as a starting point than as a motivation-driven design.
        \end{itemize}
    \item \textbf{Methodology}
        \begin{itemize}[leftmargin=*]
            \item \textbf{Hyper-SET}: provides a more rigorous formulation of energy minimization. Our energy functions are defined \textit{directly} on tokens~\eqref{eq:5} under a hyperspherical constraint. This formulation enables us to derive $\operatorname{RMSNorm}$ as a natural outcome of energy minimization in low-dimensional subspaces~\eqref{eq:7}.

            \item \textbf{ET}: defines energy over pre-normalized tokens (see Eq.(1)(6) in~\cite{hoover2024energy}) rather than tokens per se, bypassing the constrained optimization step. As a result, the role of normalization in ET is more heuristic than principled, following standard pre-norm practices rather than emerging from the underlying energy.
        \end{itemize}
    \item \textbf{Implementations}
        \begin{itemize}[leftmargin=*]
            \item \textbf{Hyper-SET}: a) applies alternating minimization that results in attention and feedforward modules \textit{sequentially}, reflecting the original Transformer structure; b) adopts adaptive, learnable step sizes conditioned on the input and iteration index~\eqref{eq:12}, allowing the model to modulate its energy descent dynamically. 
            \item \textbf{ET}: a) performs energy updates via auto-differentiation that results in a \textit{parallelized} fashion; b) uses fixed step sizes. 
        \end{itemize}
    \item \textbf{Emperical Verification}
        \begin{itemize}
            \item \textbf{Hyper-SET}: a) confirms that the designed energy decreases in the forward pass in Figures~\ref{fig:energy} and~\ref{fig:rank angle};  b) supports generalizations beyond softmax attention (\eg, different energy functions leading to alternative attention schemes), as illustrated in Tables~\ref{tab:generalized attention} and~\ref{tab:generalized feedforward}.
            \item \textbf{ET}: a) does not offer such explicit verification, leaving it unclear whether its dynamics faithfully track the energy descent objectives; b) Additionally, Hyper-SET achieves competitive performance with vanilla Transformers on tasks such as image classification and inference (\eg, Sudoku), whereas ET does not demonstrate comparable results in these domains.
        \end{itemize}
\end{itemize}
\subsection{Practical Implications and Broader Impact}
\label{section:implications and impact}

\begin{itemize}[leftmargin=*]
    \item \textbf{Why Study This Model}: This work proposes a principled approach to Transformer design by modeling representation learning as an energy minimization on the hyperspace. Unlike prior efforts such as Energy Transformer~\cite{hoover2024energy} and CRATE~\cite{yu2023white}, which either diverge from their theoretical formulations or lack generality to derive new architectures, our model directly formulates energy functions on tokens with improved rigorousness, while supporting a spectrum of alternative designs.

    \item \textbf{Why CRATE Is a Fair Baseline}: CRATE \cite{yu2023white} also pursues transparency and principled design, which shares a similar spirit with our goal, making it an appropriate baseline. While engineering-heavy ViTs may excel in benchmarks, they often involve significant redundancy. We aim to advance more compact, describable, and empirically competitive model design for next-generation architectures.
    
    \item \textbf{Interpretability}: We view the forward pass of Hyper-SET as a dynamical system. It features greater interpretability than vanilla Transformer in the sense that this dynamics is more readily describable and characterized by a meaningful quantity-the energy function-and grounded in well-established principles as maximum likelihood estimation. Beyond being merely conceptual, this dynamics is quantitatively verifiable, providing an interpretable and testable framework for understanding representation evolution.
    
    \item \textbf{A General Principle Beyond Canonical Hopfield Energy}: The Hopfield energy we employ in the main paper serves as one instantiation under the proposed general principle. Our formulation allows for broader energy-based designs—such as kernel-based alternatives to $\operatorname{logsumexp}$—enabling principled generalizations beyond standard attention mechanisms.
    
    % \item \textbf{Why This Regularizer Matters}: Our model achieves strong empirical results across tasks (e.g., CIFAR-10/100, ImageNet-100, Sudoku) while adhering strictly to a theoretically sound design. This demonstrates that structured, interpretable models can be competitive with or outperform tuned Transformer variants—showing that principled architecture design is a promising research direction.
    
    % \item \textbf{Outlook and Future Directions}: 
    % Our energy-based formulation can be extended with advanced optimization methods (e.g., momentum, Adam) in the forward pass. This lays a foundation for future models that combine interpretability, theoretical rigor, and empirical strength in a unified design paradigm. Further improvements may integrate sparse \cite{tan-etal-2023-sparse} or mixture-of-experts \cite{csordas2024moeut} techniques.
\end{itemize}

\end{document}